%% file: acl_latex.tex
\documentclass[11pt]{article}

\usepackage[preprint]{acl}

\usepackage{times}
\usepackage{latexsym}

\usepackage[T1]{fontenc}

\usepackage[utf8]{inputenc}
\usepackage{newunicodechar}
\newunicodechar{§}{\S}
\usepackage{fvextra}
\usepackage{iftex}
\ifPDFTeX
\newcommand{\reviewmapverbatimfont}{\ttfamily}
\else
\makeatletter
\newcommand{\reviewmapverbatimfont}{\font\ReviewMapVerbatimFont="[Inconsolatazi4-Regular.otf]" at \f@size pt\ReviewMapVerbatimFont}
\makeatother
\fi
\fvset{formatcom=\reviewmapverbatimfont}

\usepackage{microtype}

\usepackage{inconsolata}

\usepackage{graphicx}
\usepackage{booktabs}
\usepackage{makecell}
\usepackage{multirow}
\usepackage{xcolor}
\usepackage{colortbl}
\definecolor{ReviewMapFrameGray}{gray}{0.65}
\usepackage[most]{tcolorbox}

\title{ReviewMap: Mapping Evaluative Acts and Automation Boundaries 
in Peer-Review Comments}

\input{authors}

\begin{document}
\maketitle
\begin{abstract}
Peer-review comments check what a paper does and judge why it matters, but no existing scheme distinguishes these two evaluative acts. We argue that making these acts explicit and measuring how reliably LLMs identify them is a prerequisite for responsible LLM-assisted review analysis: humans need to inspect review judgments and their grounds. This defines a more honest mission for NLP in peer review than full automation. We present \textbf{ReviewMap}, a framework that classifies each comment as \textit{assessment} of claims and scholarly standards or \textit{calibration} of contribution value and venue fit, with finer-grained attributes applied conditionally by act type. Applied to 500 ICLR~2026 papers through multi-model LLM consensus, the act layer proves reliably automatable under the operational guideline (inter-model $\kappa \geq 0.75$), supported by five-human validation on 470 development comments, while the attribute layer requires human oversight, defining a measured automation boundary. At the corpus level, $17\%$ of reviewer--paper instances contain no panel-coded calibration, and the panel-coded calibration share is nearly $3\times$ as high in Strengths as in Weaknesses. This boundary informs three affordances: a review guideline that operationalizes the assessment-then-calibration workflow for reviewers, a per-comment rebuttal strategy scaffold for authors, and a discussion-phase dashboard for area chairs.
\end{abstract}

\section{Introduction}
\label{sec:intro}

\begin{figure}[t]
\centering
\footnotesize
\includegraphics[width=\columnwidth]{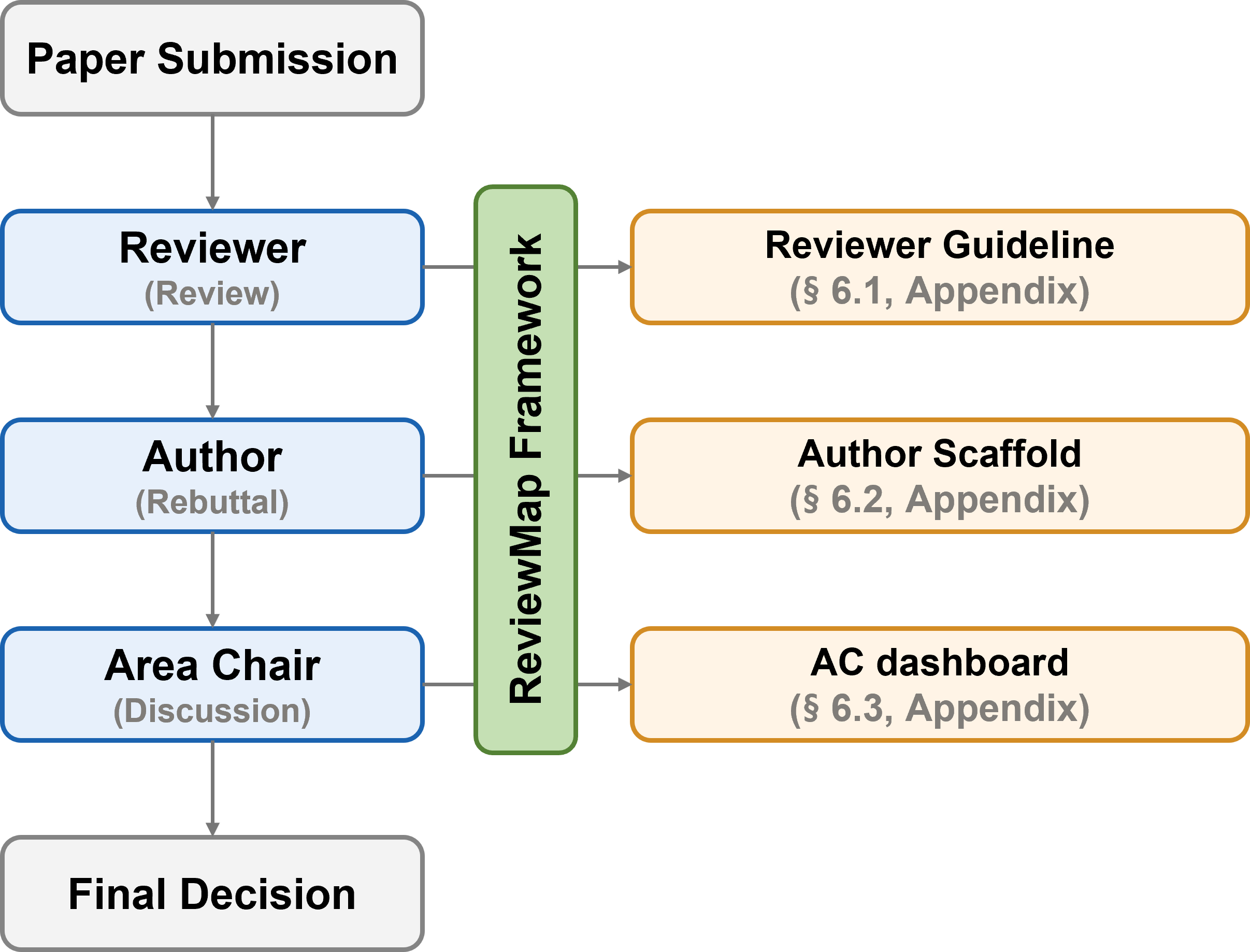}
\caption{The ReviewMap ecosystem: reviewers check assessments and ground value judgments; authors choose responses; area chairs inspect panel judgments and disagreements. These proposed affordances require users to inspect source comments and confirm attribute-based suggestions (\S\ref{sec:applications}).}
\label{fig:ecosystem}
\end{figure}

Peer review requires evaluating a submission from multiple angles: whether claims are supported by adequate evidence, whether the methodology meets the standards of the field, and whether the work constitutes a contribution significant enough for the venue. These dimensions are often recorded through separate criterion scores but ultimately inform an overall recommendation. Yet they differ in kind. Checking whether an experiment lacks a proper baseline is a different evaluative act from declaring that the contribution is too incremental. The first assesses the paper against its claims and scholarly standards; the second judges the contribution's value and venue fit. Both involve judgment; their evaluative targets and criteria differ. We call these two acts \textit{assessment} and \textit{calibration}. Venue reviewer guidelines already encode a related separation at the scoring-criterion level: ACL Rolling Review separates soundness, excitement, and overall recommendation; ICML separates soundness from significance; and NeurIPS distinguishes quality from significance. ICLR, the venue our corpus draws from, instead uses a \emph{contribution} criterion that can combine novelty, value, and venue-level significance. Across venues, however, the free-text comments interleave these evaluative acts without marking them as structurally distinct.

The growing volume of submissions has motivated NLP-for-peer-review research \citep{kuznetsov2024nlp} and raised concerns about the sustainability of the AI-venue peer-review process at scale \citep{kim2025position}. Recent work has begun deploying LLMs in this process, from detecting LLM-modified reviews \cite{pmlr-v235-liang24b} to providing automated feedback to reviewers at scale \cite{thakkar2026large} and simulating peer review dynamics with LLM agents \citep{jin-etal-2024-agentreview}. However, these efforts operate without a clear framework for determining which parts of review analysis LLMs can reliably perform and which require human judgment. Baumann et al.~(\citeyear{baumann2026stop}) argue that deploying LLMs in peer review without rigorous evaluation of their reliability risks undermining the process. Yet no existing scheme provides the measurement instrument such evaluation would require.

Prior schemes classify comments by topic, pragmatic action, or quality attributes; none distinguishes the evaluative act itself. A comment about novelty can assess a priority claim---``The approach was previously proposed by X et al.''---or calibrate contribution value---``The novelty is insufficient for this venue.'' Topic and evaluative act are orthogonal: the shared topic does not determine whether a response must establish what is new or justify why it matters.

This decomposition matters because the two acts call for different justifications. Assessments ask whether claims or standards are satisfied; calibrations ask why the contribution matters. A response to either act can use evidence and argument, but an additional ablation answers a significance judgment only when its result bears on the contribution's value. Making the act explicit helps authors address the relevant criterion and reviewers explain how assessments support an overall recommendation. Without explicit calibration, that link remains implicit; without explicit grounding, its basis is harder to engage with.

\paragraph{Our position.}
Operationalizing this distinction and measuring how reliably LLMs classify it is a prerequisite for responsible LLM-assisted analysis of review judgments. The aim is to make review judgments and uncertainty in their coding inspectable by humans. Three contributions support this position: (1)~\textbf{An act-based framework.} ReviewMap operationalizes assessment and calibration at the comment level, with act-conditional attributes. (2)~\textbf{A measured automation boundary.} Three-model consensus over $500$ ICLR~2026 papers ($17{,}654$ comments), supported by five-human development-set validation, establishes reliable act classification under the protocol and exposes lower attribute agreement. (3)~\textbf{Design implications across roles.} We articulate design implications for a reviewer guideline, an author scaffold, and an AC dashboard using this distinction and measured boundary (Figure~\ref{fig:ecosystem}).

\section{Related Work}
\label{sec:background}

Table~\ref{tab:comparison} situates ReviewMap against prior peer-review comment analysis schemes, grouped by what they classify; we walk through each tradition below.

\begin{table*}[t]
\centering\small
\setlength{\tabcolsep}{3.1pt}
\begin{tabular}{llllcc}
\toprule
Paper & Venue & Unit & Classifies & \makecell{LLM\\annot.} & Audience \\
\midrule
PeerRead \citeyearpar{kang-etal-2018-dataset} & NAACL '18 & Review & Aspect scores & No & -- \\
ASAP-Review \citeyearpar{yuan2022can} & JAIR '22 & Text segment & Aspect labels & No & R \\
Lu et al.\ \citeyearpar{lu-etal-2025-identifying} & EMNLP F'25 & Sentence & Aspect (coarse/fine) & Single & R \\
\midrule
AMPERE \citeyearpar{hua-etal-2019-argument} & NAACL '19 & Text segment & Proposition type & No & -- \\
DISAPERE \citeyearpar{kennard-etal-2022-disapere} & NAACL '22 & Sentence & Review-Action (+attributes) & No & C \\
NLPeer \citeyearpar{dycke-etal-2023-nlpeer} & ACL '23 & Sentence & Pragmatic tag & No & R \\
\midrule
SubstanReview \citeyearpar{guo-etal-2023-automatic} & EMNLP F'23 & Text segment & Substantiation & No & R/C \\
PolitePEER \citeyearpar{bharti2024politepeer} & LRE '24 & Sentence & Politeness (5 levels) & No & R/C \\
LazyReview \citeyearpar{purkayastha-etal-2025-lazyreview} & ACL '25 & Text segment & Lazy thinking & Single & R \\
Kuznetsov et al.~(\citeyear{kuznetsov2026decline}) & EMNLP '26 & Text segment / review & Quality metrics & Single & A/C \\
\midrule
\textbf{ReviewMap} (ours) &---& \textbf{Text segment} & \textbf{Evaluative act (ASM/CAL)} & \textbf{Multi} & \textbf{R/A/C} \\
\bottomrule
\end{tabular}
\caption{Selected review-text tasks; prior schemes do not distinguish assessment from calibration. Text segments cover study-specific spans, propositions, items, and comments, with differing boundaries (LazyReview: 1--5 sentences). LLM annotation counts generative models assigning the listed labels/scores, excluding models used only for preprocessing or performance comparison; Multi denotes consensus. Audience identifies roles explicitly addressed in task or use-case design: R = reviewer, A = author, C = chairs/editors; -- = unspecified.}
\label{tab:comparison}
\end{table*}

\paragraph{Aspect classification.}
PeerRead \cite{kang-etal-2018-dataset} introduced review-level aspect scores; ASAP-Review \cite{yuan2022can} extended this to span-level aspect and polarity annotation; Lu et al.~(\citeyear{lu-etal-2025-identifying}) proposed a bottom-up, LLM-driven taxonomy with 16 coarse and 39 fine-grained aspect categories. These schemes classify what a comment is \emph{about}. ReviewMap classifies what a comment \emph{does}.

\paragraph{Pragmatic action.}
AMPERE \cite{hua-etal-2019-argument} introduced proposition-level labels for review argumentation. DISAPERE \cite{kennard-etal-2022-disapere} synthesized prior schemes into sentence-level properties including Review-Action, where ``Evaluative'' is a single category. NLPeer \cite{dycke-etal-2023-nlpeer} contributed a multi-domain ethically licensed corpus with pragmatic tags. These schemes classify \emph{which discourse role} a comment plays (evaluate, request, recap, \dots); DISAPERE is our closest structural precedent. ReviewMap operates one layer deeper: we decompose DISAPERE's single ``Evaluative'' role into assessment and calibration, and replace human annotation with multi-model LLM consensus for scalability.

\paragraph{Review quality attributes.}
SubstanReview \cite{guo-etal-2023-automatic} measures claim substantiation in 550 reviews, reporting a declining substantiation trend across NLP conferences over time. PolitePEER \citep{bharti2024politepeer} annotates peer-review comments by politeness intensity across five levels. LazyReview \cite{purkayastha-etal-2025-lazyreview} derives an 18-category annotation scheme from ARR guidelines and validates it through a controlled experiment. These schemes assess review quality without distinguishing assessment from calibration. \citet{kuznetsov2026decline} score weaknesses and related review items for actionability and grounding alongside review-level quality. \citet{amirshahi2026cover} find LLM-based review metrics sensitive to meaning-preserving rewrites. ReviewMap first separates evaluative acts, then applies act-specific attributes.

\paragraph{Multi-model consensus.}
Single-model LLM annotation is effective but contested in reliability \cite{gilardi2023chatgpt,bavaresco-etal-2025-llms}. Verga et al.~(\citeyear{verga2024replacing}) show that a panel of diverse LLMs outperforms any single judge for generation evaluation. Courvoisier et al.~(\citeyear{courvoisier2026beyond}) apply multi-model agreement to classification and extraction tasks, introducing a withheld-judgment protocol for cases without consensus. Wang et al.~(\citeyear{wang2023selfconsistency}) provide the theoretical basis for self-consistency sampling. We extend panel consensus to multi-layer subjective annotation; reliability stratifies by layer.

\section{The ReviewMap Framework}
\label{sec:framework}

\textbf{ReviewMap} separates two \emph{evaluative acts}: assessment of claims and scholarly standards, and calibration of contribution value and venue fit. We choose this distinction because whether a claim is supported or a standard met and whether a contribution is valuable ask different evaluative questions. Level~1 classifies the act (§\ref{sec:framework:level1}); conditional Level~2 attributes expose the assessment's criteria (P1/P2) or the calibration's stated grounds (§\ref{sec:framework:level2}), so readers can inspect the basis of each judgment.

\subsection{Level~1: Evaluative Act (ASM / CAL)}
\label{sec:framework:level1}

Each atomic comment in our corpus is assigned exactly one of three act codes (Table~\ref{tab:framework}, top). The boundary between \textsc{Asm} and \textsc{Cal} is the locus of \emph{value commitment}: \textsc{Asm} evaluates support for claims and conformity with scholarly standards, while \textsc{Cal} judges the contribution\textquotesingle s novelty, significance, or venue fit. Both can be positive or negative and involve interpretation. Questions are classified by their evaluative function; mixed comments receive their dominant act. \textsc{Flag} is the fallback when neither act can be assigned under the guideline (Appendix~\ref{app:guideline}).

\begin{table}[t]
\centering\footnotesize
\setlength{\tabcolsep}{5pt}
\begin{tabular}{@{}p{0.25\columnwidth} p{0.31\columnwidth} p{0.30\columnwidth}@{}}
\toprule
Code / Attribute & Definition & Example \\
\midrule
\rowcolor{gray!10} \multicolumn{3}{@{}l}{\textsc{Level 1 --- Evaluative Act}} \\
\addlinespace[2pt]
\textbf{ASM}\newline(assessment) & Evaluates claims, methods, or scholarly standards. & Ablations don't report variance \\
\textbf{CAL}\newline(calibration) & Commits to a judgment about the paper's significance. & The empirical gains are marginal \\
\textbf{FLAG} & ASM/CAL cannot be assigned under the guideline. & See major weaknesses above \\
\midrule
\rowcolor{gray!10} \multicolumn{3}{@{}l}{\textsc{Level 2 --- Act-conditional attributes}} \\
\addlinespace[2pt]
claim\_relevant\newline(P1, for ASM) & Evaluates against a paper-specific claim. & Paper claims A, but Table shows B. \\
standard\_\allowbreak relevant\newline(P2, for ASM) & Evaluates against a field-wide standard. & Ablations should report variance. \\
grounded\newline(for CAL) & Value judgment anchored in explicit evidence. & Novelty is limited; the derivation resembles prior work. \\
\bottomrule
\end{tabular}
\caption{The ReviewMap framework. Level~1 classifies each comment by its evaluative act; Level~2 attaches binary attributes conditioned on the Level~1 act. Full operational definitions in Appendix~\ref{app:guideline}; corpus distribution and reliability in §\ref{sec:results}.}
\label{tab:framework}
\end{table}

\subsection{Level~2: Act-conditional attributes}
\label{sec:framework:level2}
The second layer attaches binary attributes conditioned on the Level~1 act (Table~\ref{tab:framework}, bottom). For \textsc{Asm}, two attributes record what the assessment evaluates against: a specific claim made by the paper (claim\_relevant, P1) and/or a field-wide standard or convention (standard\_relevant, P2). For \textsc{Cal}, a single attribute records whether the reviewer's value judgment is anchored in explicit evidence (grounded), following the substantiation literature \citep{guo-etal-2023-automatic}. \textsc{Flag} comments carry no Level~2 attributes.

P1 and P2 are \emph{orthogonal dimensions}: a comment can score yes on both when a specific claim also conflicts with a field standard. P2 and \emph{grounded} target adjacent but orthogonal axes: P2 is a \emph{targeting} property (does the comment invoke a shared community standard?), while \emph{grounded} is a \emph{justification} property (does the comment supply explicit evidence for a personal judgment?).

\paragraph{From a comment to a coding.}
Two positive corpus comments illustrate the distinction. ``The claims are supported with theoretical justification'' assesses claims: all three model labels are ASM, P1=yes, P2=no. ``The idea is very simple but impactful'' judges contribution value: all three are CAL, grounded=no, with no explicit support for ``impactful.'' Thus act differs from sentiment. Labels are aggregated first across runs, then across models (§\ref{sec:method:consensus}); Appendix~\ref{app:actual-prompt} links source identifiers, inputs, and output format.

\section{Methodology}
\label{sec:method}

\subsection{Data collection}
\label{sec:method:corpus}

We sample 500 ICLR~2026 submissions from OpenReview \citep{soergel2013openreview}, stratified by 21 primary areas crossed with accept/reject decision (seed 42; $38.8\%$ accepted; at least three accepted and three rejected papers per area). Sampling across areas examines the framework's coverage beyond a single subfield.

Each review is segmented into comments by a rule-based parser that uses list markers and paragraph boundaries within the venue's Strengths / Weaknesses / Questions sections. The artifact preserves the batch-specific preprocessing profiles. The 500 papers yield $17{,}654$ comments across $1{,}955$ reviewer--paper instances. This is the unit of annotation; everything downstream (§\ref{sec:results}, §\ref{sec:applications}) operates over these comments rather than over full reviews.

Separately, before the main run, we compiled an additional development set (12 papers, 470 comments). The dev set supports the open-weight, frontier-LLM, and human comparisons in §\ref{sec:method:validation}.

\subsection{Annotation design}
\label{sec:method:pipeline}

We developed a single annotation guideline applied uniformly across all models (see Appendix~\ref{app:guideline} for the full text and worked examples; §\ref{sec:framework} summarizes the operational definitions). Each prompt joins a single paper's segmented comments with the guideline; the LLM returns one structured row per comment with the Level-1 code and the applicable Level-2 attributes.

\paragraph{Annotation inputs.}
Each per-paper prompt contains the title, abstract, keywords, reviewer identifiers, scores and confidence, section headers, comment identifiers and text, together with the shared guideline. The wrapper extracts claims from the supplied abstract; the full paper is not supplied. Human dev-set reading sheets also include scores and confidence. We evaluate the annotation pipeline under this input condition.

\paragraph{Inference.}
We use vLLM \citep{kwon2023efficient} at temperature 0.6 for self-consistency sampling \citep{wang2023selfconsistency}. The 3-model $\times$ 3-run pipeline on the 500-paper main set consumes approximately $350$ GPU-hours across two hardware configurations; the released artifact contains the archived prompts, parsed model outputs, and aggregation code (Appendix~\ref{app:reproducibility}; hardware accounting is in Table~\ref{tab:gpu-hours}).

\subsection{Multi-model consensus design}
\label{sec:method:consensus}

We select three open-weight LLMs from different groups and architecture families: Qwen 3.5~\cite{qwen3-5_2026}, Gemma 4~\cite{gemma4_2026}, and Nemotron 3-Super~\cite{nemotron3_2026} (Table~\ref{tab:models}).

\begin{table}[t]
\centering\footnotesize
\begin{tabular}{l l l}
\toprule
Model & Group & Active params \\
\midrule
Qwen 3.5 & Alibaba & 27\,B (dense) \\
Gemma 4 & Google & 31\,B (dense) \\
Nemotron 3-Super & NVIDIA & 12\,B / 120\,B (MoE) \\
\bottomrule
\end{tabular}
\caption{The three open-weight LLMs used for the corpus annotation, spanning three research groups.}
\label{tab:models}
\end{table}

\begin{figure*}[t]
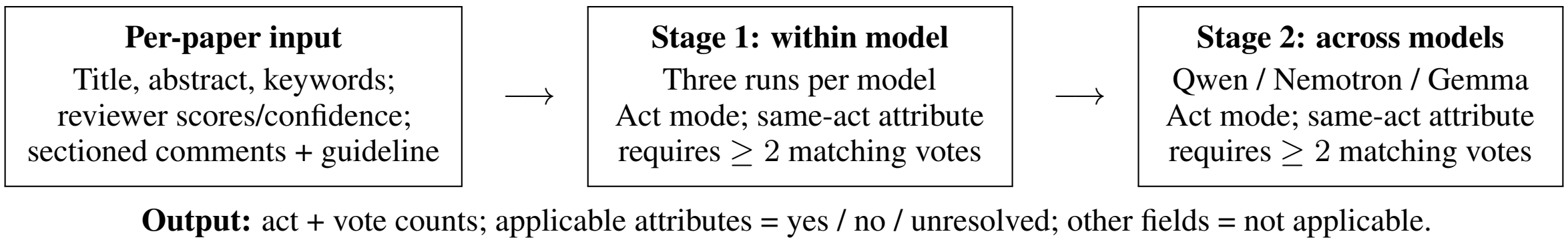

\centering\small
\setlength{\fboxsep}{5pt}
\fbox{\begin{minipage}[c]{0.26\textwidth}\centering
\textbf{Per-paper input}\\[2pt]
Title, abstract, keywords;\\
reviewer scores/confidence;\\
sectioned comments + guideline
\end{minipage}}
\quad$\longrightarrow$\quad
\fbox{\begin{minipage}[c]{0.24\textwidth}\centering
\textbf{Stage 1: within model}\\[2pt]
Three runs per model\\
Act mode; same-act attribute\\
requires $\geq 2$ matching votes
\end{minipage}}
\quad$\longrightarrow$\quad
\fbox{\begin{minipage}[c]{0.24\textwidth}\centering
\textbf{Stage 2: across models}\\[2pt]
Qwen / Nemotron / Gemma\\
Act mode; same-act attribute\\
requires $\geq 2$ matching votes
\end{minipage}}
\par\vspace{5pt}
\textbf{Output:} act + vote counts; applicable attributes = yes / no / unresolved; other fields = not applicable.
\caption{The two-stage voting pipeline. The main corpus contains $17{,}654$ comments from $500$ papers. The same act-conditioned attribute rule applies at both stages. Act ties follow fixed orders and remain visible in vote metadata; unresolved attributes identify insufficient support for human inspection.}
\label{fig:pipeline}
\end{figure*}

The annotation pipeline runs each model \emph{three times} and resolves to a per-model coding via intra-model voting (Stage~1). The three per-model codings are then resolved to a final consensus via inter-model voting (Stage~2). This exposes run-to-run variation within models and disagreement between model families (Figure~\ref{fig:pipeline}).

\paragraph{Act and attribute voting.}
At each stage, the act is the mode of available codes. Ties retain the first available run (Stage~1) or the first model in the fixed Qwen--Nemotron--Gemma order (Stage~2), with the vote counts preserved. Missing outputs are excluded, not treated as \textsc{Flag}. For each applicable Level-2 field, only voters whose act matches the selected act contribute a yes/no vote, and \emph{at least two agreeing votes are required at each stage}. Thus a single valid attribute cannot determine a model or final label. A tie or fewer than two supporting votes yields \texttt{unresolved}; fields outside the selected act are \texttt{not\_applicable}. These distinct states replace the raw output's ambiguous NA. Table~\ref{tab:attribute-tiers} distinguishes three supporting votes, two with dissent, two without a third valid vote, and unresolved attribute fields. Appendix~\ref{app:evidence} gives full conventions and the retained joint-schema exceptions.

Unlike prior schemes that train a dedicated classifier on labeled examples \citep{kennard-etal-2022-disapere, purkayastha-etal-2025-lazyreview}, ReviewMap uses LLMs directly as annotators against a written guideline, eliminating the train/test split and making schema iteration possible without retraining.

\subsection{Pipeline validation on the dev set}
\label{sec:method:validation}

We evaluate dev-set act labels against human and frontier-model references before interpreting the main-corpus analysis.

\paragraph{Five-human reference.}
An ML-expert author and four additional ML graduate annotators each code all 470 comments under the same guideline, without model outputs, one another's labels, or adjudication. This comparison evaluates Level~1 only. We use the human mode as a comparison reference, not an error-free gold standard; the single 2--2--1 tie follows ASM $>$ CAL $>$ FLAG. Human agreement is Fleiss' $\kappa=.551$ (mean pairwise Cohen's $\kappa=.555$, range $.468$--$.649$; raw agreement $88.1\%$). All five agree on $352/470$ comments; $102$ of the remaining $118$ involve ASM/CAL-only disagreement.

\paragraph{Frontier-LLM anchor.}
We additionally collect a three-model frontier consensus via manual chat-UI sessions: GPT-5.5 (ChatGPT) \citep{openai2026gpt55}, Gemini~3.1 Pro \citep{google2025gemini3}, and Claude Opus~4.6 \citep{anthropic2026claude46}, each run three times per paper (108 sessions). Their Level-1 labels use the same two-stage mode procedure, with GPT--Gemini--Opus as the panel tie order. Appendix~\ref{app:evidence} reports the dev output audit and individual-model comparisons.

\paragraph{Validation result.}
Both panels agree substantially with the five-human reference under the Cohen's $\kappa$ \citep{cohen1960coefficient} interpretation of \citet{landis1977measurement}: $.633$ for open-weight and $.686$ for frontier models (Table~\ref{tab:human-validation}). Frontier--open-weight agreement is $.836$. Across five individual annotators and both panels, pairwise $\kappa$ ranges from $.505$ to $.658$ (mean $.581$), overlapping the human--human range. Agreement approaches $97\%$ on human-unanimous items and decreases on contested items, locating the remaining ambiguity. CAL precision/recall is $.560/.824$ for the open-weight panel and $.613/.902$ for the frontier panel; both predict $75$ CALs versus $51$ in the human reference. These results support act coding under the evaluated protocol. Open-weight panel agreement with the human reference is $95.8\%$ on model-unanimous items and $56.5\%$ otherwise (Table~\ref{tab:dev-vote-inspection}), motivating visible dissent for inspection.

\begin{table}[t]
\centering\footnotesize
\setlength{\tabcolsep}{4pt}
\begin{tabular}{@{}lrr@{}}
\toprule
Human-reference comparison & Open-weight & Frontier \\
\midrule
All 470: Cohen's $\kappa$ & .633 & .686 \\
\quad raw agreement & 90.0\% & 91.5\% \\
Human-unanimous ($n=352$) & 96.6\% & 97.2\% \\
Human-contested ($n=118$) & 70.3\% & 74.6\% \\
\bottomrule
\end{tabular}
\caption{Level-1 dev validation against the five-human mode. All five annotators label all 470 items; subset rows report raw agreement for human-unanimous and contested items.}
\label{tab:human-validation}
\end{table}

\subsection{Scaling to the 500-paper main set}
\label{sec:method:scaling}

We apply the three-model pipeline to all 500 papers. All $17{,}654$ expected comments receive a consensus act. The final applicable-attribute coverage is $96.26\%$ for P1, $95.95\%$ for P2, and $91.42\%$ for grounding; $1{,}115$ comments retain at least one unresolved applicable field.

Only generated rows whose (paper, reviewer, comment\_id) occurs in the expected segmented corpus enter aggregation. Duplicate keys retain the last row, matching the saved-output loader; missing and extra keys are audited. Raw generation errors and deliberate aggregation abstentions are reported separately in \S\ref{sec:results:failure}.

\paragraph{Agreement measurement.}
Intra-model agreement uses Fleiss' $\kappa$ over three runs; inter-model agreement uses pairwise Cohen's $\kappa$ over model labels. Compared raters must all emit valid acts; attribute coefficients additionally require the same applicable act and binary values. Appendix~\ref{app:evidence} reports $n$, act-eligible coverage, observed agreement, and matched-item sensitivity. These coefficients measure label consistency under the annotation protocol.

\section{Results}
\label{sec:results}

\subsection{What ReviewMap reveals}
\label{sec:results:corpus}

\paragraph{Act-layer classification is reliable.}
The Level-1 act code is reliably classified by all three open-weight LLMs (Table~\ref{tab:function-kappa}). Intra-model 3-run self-consistency $\kappa$ is Substantial-to-Almost-Perfect for every model; inter-model pairwise $\kappa$ is uniformly Substantial across all three model pairs. Distributions converge tightly---\textsc{Asm} $80.5\%$ / \textsc{Cal} $17.4\%$ / \textsc{Flag} $2.2\%$---and the three-way consensus achieves \textbf{$88.4\%$ unanimous, $11.3\%$ 2-of-3, $0.3\%$ no-majority}.

\begin{table}[t]
\centering\footnotesize
\setlength{\tabcolsep}{4pt}
\begin{tabular}{@{}ll@{}}
\toprule
Metric & Value \\
\midrule
Intra-Fleiss $\kappa$ (Q\,/\,G\,/\,N)                                   & 0.84 / 0.96 / 0.78 \\
Inter-Cohen $\kappa$ (QG / QN / GN) & 0.77 / 0.77 / 0.75 \\
Act distribution \% (ASM/CAL/FLAG)                                & 80.5 / 17.4 / 2.2 \\
Consensus split \% (unan./2-of-3/no-maj)                          & 88.4 / 11.3 / 0.3 \\
\bottomrule
\end{tabular}
\caption{Act-layer reliability across the three open-weight models (main set, $n=17{,}654$ comments). Both intra- and inter-model agreement are Substantial or higher. Intra-model $n$ is $17{,}654/17{,}653/17{,}642$ for Q/G/N; inter-model $n=17{,}654$ throughout.}
\label{tab:function-kappa}
\end{table}

\begin{figure}[t]
\centering
\includegraphics[width=\columnwidth]{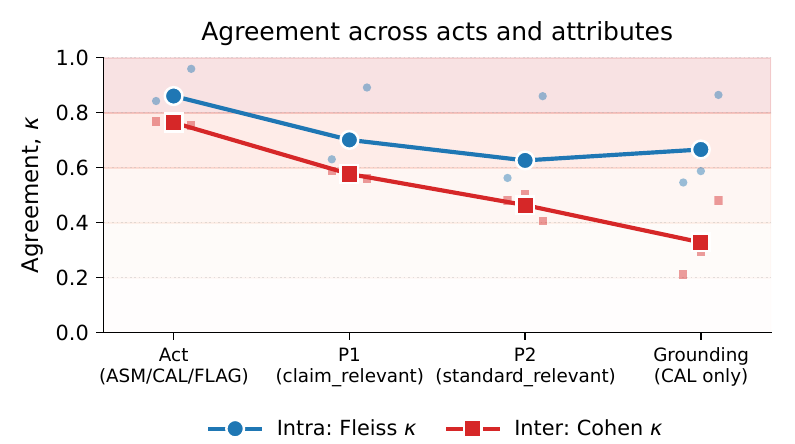}
\caption{Agreement across acts and attributes. Intra-model points use Fleiss' $\kappa$ over three runs; inter-model points use Cohen's $\kappa$ over pairs of model labels. Mean inter-model agreement decreases across act, P1, P2, and grounding. Attribute samples are conditional on matching applicable acts and binary labels; denominators and observed agreement appear in Appendix~\ref{app:evidence}.}
\label{fig:attribute-agreement}
\end{figure}

\paragraph{Agreement is lower for attributes.}
\label{sec:results:attribute}
The Level-2 attributes (P1\,/\,P2\,/\,grounded) tell a different story. Mean inter-model $\kappa$ decreases from $.764$ at the act layer to $.577$, $.464$, and $.328$, respectively (Figure~\ref{fig:attribute-agreement}). The trend varies across model pairs: Nemotron--Gemma rises from $.407$ on P2 to $.480$ on grounding. Intra-model Level-2 $\kappa$ remains Almost Perfect for Gemma ($.86$--$.89$), is Moderate-to-Substantial for Qwen ($.55$--$.63$), and Moderate for Nemotron ($.46$--$.59$). For grounding, within-model agreement exceeds between-model agreement in every pair even on matched items (Appendix~\ref{app:evidence}), consistent with model-specific decision thresholds. The attribute layer therefore structures response options that require human confirmation (\S\ref{sec:apps:caveats}). Three-model unanimity is about $60\%$ for P1/P2 and $37.8\%$ for grounding, while unresolved attribute fields increase from $3.7$--$4.0\%$ to $8.6\%$ (Table~\ref{tab:attribute-tiers}).

\begin{table}[t]
\centering\footnotesize
\setlength{\tabcolsep}{3pt}
\begin{tabular}{@{}lrrrrr@{}}
\toprule
Field & $n$ & 3/3 & 2/3 & 2/2 & Unres. \\
\midrule
P1 (ASM) & 14{,}203 & 59.46 & 27.49 & 9.31 & 3.74 \\
P2 (ASM) & 14{,}203 & 60.87 & 27.42 & 7.66 & 4.05 \\
Grounded (CAL) & 3{,}067 & 37.82 & 35.31 & 18.29 & 8.58 \\
\bottomrule
\end{tabular}
\caption{Final attribute support tiers (\% of applicable $n$). 3/3 = three agreeing valid votes; 2/3 = two agreeing with one dissent; 2/2 = two agreeing with no third valid, same-act vote. Unres. = tie or insufficient valid support.}
\label{tab:attribute-tiers}
\end{table}

\begin{table}[t]
\centering\footnotesize
\setlength{\tabcolsep}{4pt}
\begin{tabular}{@{}ll@{}}
\toprule
Observation & Value \\
\midrule
Act distribution (ASM / CAL / FLAG)              & 80.5 / 17.4 / 2.2 \% \\
Zero-CAL reviewer--paper instances & $333/1{,}955$ ($17.0\%$) \\
Per-reviewer CAL (median / mean)   & $1$ / $1.57$ \\
\quad reviewers with $\leq 2$ CALs & $81.4\%$ \\
\midrule
\multicolumn{2}{@{}l}{\emph{ASM\,/\,CAL\,/\,FLAG share by section (\%)}} \\
\quad Strengths   & 65.1 / 34.5 / 0.4 \\
\quad Weaknesses  & 85.5 / 12.2 / 2.2 \\
\quad Questions   & 89.2 / 6.9 / 3.9 \\
\bottomrule
\end{tabular}
\caption{Corpus-level findings from final three-model consensus acts. Section shares use $5{,}269$ Strengths, $7{,}390$ Weaknesses, and $4{,}995$ Questions comments; review statistics use $1{,}955$ reviewer--paper instances.}
\label{tab:corpus-findings}
\end{table}

\paragraph{Explicit calibration is sparse in review text.}
The panel classifies $17.4\%$ of comments as \textsc{Cal} (explicit value commitments), and the remainder assess claims or scholarly standards, or fall outside these two acts. At the review level, \textbf{17.0\%} of reviewer--paper instances contain \emph{no} panel-coded \textsc{Cal}; the median reviewer issues a single \textsc{Cal} per paper (mean $1.57$; $81.4\%$ issue $\leq 2$). Across 21 primary areas ($11$--$54$ papers each), comment-level CAL shares range from $12.4\%$ to $21.7\%$ (Table~\ref{tab:area-calibration}), describing variation within ICLR rather than cross-venue generalization.

\paragraph{Strengths are calibration-heavy; Weaknesses emphasize assessment.}
Where \textsc{Cal} does appear, it concentrates in Strengths: $34.5\%$ of Strengths comments are \textsc{Cal} versus $12.2\%$ in Weaknesses and only $6.9\%$ in Questions---a nearly $3\times$ gap between Strengths and Weaknesses (Table~\ref{tab:corpus-findings}). Section headers alone do not identify evaluative acts: even Strengths sections are $65.1\%$ \textsc{Asm} (most ``strengths'' assess claims or scholarly standards), and Weaknesses sections are $85.5\%$ \textsc{Asm} (reviewers tend to assess a weakness against those criteria rather than explicitly judge its bearing on the contribution's value). These panel codings suggest that reviewers express value judgments more often when complimenting than when critiquing. On dev, the human mode labels $1/189$ Weaknesses comments as CAL versus $14/189$ for the open-weight panel; both show higher CAL shares in Strengths, while the magnitude of the corpus contrast remains panel-dependent (Appendix~\ref{app:evidence}).

These four observations motivate §\ref{sec:applications}: the act layer is reliably automatable but the attribute layer requires care, and calibration is both sparse and section-asymmetric. Authors, reviewers, and area chairs all need a view of \emph{where} value commitments live, not just \emph{how many} comments a section contains---the basis for the audience-specific affordances.

\subsection{Per-model failure analysis}
\label{sec:results:failure}

Table~\ref{tab:failure-modes} audits missing/extra IDs, duplicates, and applicable attributes emitted as NA. Across runs, Qwen omits P2 more than P1 ($313$ versus $84$ applicable NAs); Gemma shows the reverse ($1{,}454$ P1 versus $520$ P2). Generation failures are distinct from the two-vote rule's unresolved outputs.

Aggregation covers every expected act. Of all comments, $16{,}155/17{,}654$ ($91.51\%$) are non-\textsc{Flag} with all applicable attributes resolved. Independent attribute-field voting leaves $40/14{,}203$ ASM comments with P1=P2=no, violating the guideline's at-least-one-yes constraint. The released labels retain these joint-schema exceptions alongside resolved-field coverage.

\section{Design Implications}
\label{sec:applications}

The measured boundary in \S\ref{sec:results} connects three inspection tasks: reviewers examine their evaluative targets and grounds, authors identify what responses must justify, and area chairs trace what remains disputed. The sketches retain model votes and require human confirmation of attribute-derived suggestions. Appendix~\ref{app:dashboard} follows a concern from contested act coding to a manually interpreted response, connecting these role-specific questions.

\subsection{For reviewers: assessment-then-calibration workflow}
\label{sec:apps:reviewer}

The framework provides a structural self-check: does the review explicitly evaluate contribution value alongside claims and scholarly standards? In our corpus, $17\%$ of reviewer--paper instances contain no consensus-coded calibration in the sampled Strengths, Weaknesses, and Questions comments. This invites inspection of the explicit links between assessments and value judgments. Our practical reviewer guideline (Appendix~\ref{app:reviewer-guide}) maps assessment to \emph{Claim--Evidence Alignment} and \emph{Scholarly Standard Assessment}, and calibration to \emph{ground your judgment}. The corpus norms in \S\ref{sec:results:corpus} (median 1 calibration per review; $81.4\%$ of reviewers issue $\leq 2$) provide descriptive context for that inspection.

\subsection{For authors: what a response must justify}
\label{sec:apps:author}

The distinction identifies what a response must justify. Assessment comments call for addressing claim support or scholarly standards; calibration comments call for explaining the contribution's significance or contextualizing its value. Either response may combine evidence and argument. The coding routes authors toward five candidate strategies (\textsc{Leverage}, \textsc{Defend}, \textsc{Concede}, \textsc{Accept}, \textsc{Clarify}); the author selects and develops the response. Appendix~\ref{app:author-scaffold} specifies the mapping and quantifies its conditional attribute dependence and unresolved cases. This is a prospective decision-support proposal; large-scale validation against author-side rebuttal corpora, complementing author-input-driven response generation \citep{ruan2026authorinloop}, remains future work.

\subsection{For area chairs: discussion-phase panel view}
\label{sec:apps:ac}

The same distinction lets area chairs inspect how a panel assesses the paper and judges its contribution, beyond its raw scores. \citet{wang2026evidenceconflicts} weight conflicting opinions by support and review quality to generate meta-reviews; our proposed act-based view supports human inspection. An area chair can see which reviews explicitly calibrate contribution value and which focus on claims or standards, where calibrations cluster across the panel, and which comments the LLM panel itself disagreed on (the $11.3\%$ with 2-of-3 consensus warrant closer reading). A discussion-phase dashboard in Appendix~\ref{app:dashboard} traces two concerns in a polarized four-reviewer panel ($2/4/6/8$). The two-stage aggregation marks one concern as grounded and the other as unresolved; both remain available for the AC to inspect alongside manually linked author responses.

\subsection{Responsible use}
\label{sec:apps:caveats}

Each affordance above relies on the act layer, where LLM consensus is reliable under the measured conditions. Attribute-level outputs are advisory and require human confirmation. Any user-facing implementation should display the source comment, individual model votes, and agreement tier; mark 2-of-3 acts as contested; and treat three-way act disagreements as unresolved despite the corpus's deterministic tie-break. Unresolved attributes remain unresolved, and every attribute-derived menu or filter requires confirmation, including resolved values with dissenting votes. Users must be able to override codings. These sketches do not generate reviews, score reviewers, rank papers, or recommend decisions. Making the measured boundary visible is the contribution: stakeholders can inspect and revise the classifications supporting their judgments.

\paragraph{Alternative views.}
Criterion scores separate soundness and significance but do not identify the judgments or grounds expressed in individual comments. Aspect and pragmatic labels can complement act coding: the same topic or request may concern claim support or contribution value, requiring different justifications. Better models or guidelines may reduce disagreement; the measured boundary is conditional. Agreement about which act a comment expresses does not require agreement about the paper's merit. Our position is that NLP support should expose evaluative targets, grounds, and coding uncertainty for human deliberation.

\section{Conclusion}
\label{sec:discussion}

ReviewMap makes assessment of claims and scholarly standards and calibration of contribution value explicit at the comment level. Across $500$ papers and $17{,}654$ comments, its codings reveal sparse explicit calibration ($17\%$ zero-CAL reviewer--paper instances) and section asymmetry (a nearly $3\times$ higher CAL share in Strengths than Weaknesses). The measured reliability boundary and five-human validation support using act classification to structure inspection, while finer attributes remain advisory. The reviewer guideline, author scaffold, and AC dashboard translate that boundary across roles. We argue that exposing \emph{what LLMs can reliably structure and where human judgment must remain explicit} defines a more honest mission for NLP in peer review than full automation, complementing the rigor call of \citet{baumann2026stop}.

\section*{Limitations}
\label{sec:limitations}

Our empirical analysis covers English reviews from ICLR~2026, with Level-1 human validation on 470 development comments. The $17\%$ zero-CAL finding (§\ref{sec:results}) concerns explicit calibration in the sampled review sections; it provides no direct evidence about reviewers' unstated reasoning or the validity of their recommendations. CAL overprediction relative to the dev human reference may affect both comment-level prevalence and zero-CAL counts; the dev comparison cannot calibrate main-corpus estimates. Generalization to other venues remains to be evaluated. The Yes--Yes--Yes donation protocol in NLPeer \citep{dycke-etal-2023-nlpeer} and cross-venue corpora such as Re2 \citep{zhang2025re2} offer precedents for extending this analysis with appropriate data access.

The five-human study supports act classification under the operational guideline, with interpretive disagreement on 118 of 470 comments. Human-majority labels therefore provide an imperfect reference, and agreement on the unanimous subset should be interpreted alongside the full-set results. Human validation is limited to Level~1. Attribute agreement is conditional on shared acts and available binary values, with different samples and prevalences across fields (§\ref{sec:results}). The observed agreement differences are specific to this protocol. Annotation accuracy, intrinsic subjectivity, and shared model bias require separate evidence. The guideline's traceable-evidence criterion and the wrapper's reference-to-assessment wording differ for grounding (Appendix~\ref{app:actual-prompt}); their contribution to disagreement remains untested. Sensitivity to alternative guidelines, a concern in LLM annotation \citep{casanova2026adaptability}, also remains to be evaluated.

The three affordances illustrate potential uses of the framework, including a worked dashboard case (Appendix~\ref{app:dashboard}); their effects on human decisions require further study. Rule-based segmentation can yield fragmentary comments, and the attributes capture different aspects of review text: P1/P2 identify assessment targets, while grounding identifies stated support for value judgments. These annotations describe review structure rather than verify the correctness of an assessment or the merit of a paper. Their use in decision support retains the human confirmation described in §\ref{sec:apps:caveats}.

\section*{Ethical considerations}
\label{sec:ethics}

The source papers and reviews are publicly available ICLR~2026 submissions on OpenReview, following prior work on public peer-review corpora \citep{kennard-etal-2022-disapere, darcy-etal-2024-aries, dycke-etal-2023-nlpeer}. The five-human study labels these existing comments at Level~1, with anonymized annotator identifiers. The four additional annotators were ML graduate students recruited through personal contacts in neighboring laboratories. Participation was voluntary and unpaid, with acknowledgment promised upon acceptance. They agreed to research use and public sharing of their annotations and received the annotation guideline without additional task training. No institutional ethics review was sought for this study. The released artifact retains paper identifiers and the venue's public pseudonymous reviewer identifiers (e.g., \texttt{Reviewer\_NydX}) unchanged, matching the DISAPERE / ARIES convention; we do not perform additional pseudonymization and do not attempt to de-anonymize reviewers. The case study in Appendix~\ref{app:dashboard} quotes short illustrative fragments and uses short forms of the venue-assigned reviewer pseudonyms (e.g., \texttt{LnWH}) for readability; full review and rebuttal threads remain accessible at the OpenReview source URLs.

\paragraph{Risks and intended-use scope.}
ReviewMap is intentionally a classification framework, not a generation system: the released code does not synthesize review or rebuttal text. We share the position of \citet{dycke-etal-2023-nlpeer} and \citet{baumann2026stop} that automatic review generation is a category of NLP application where benefits are unclear and risks to academic quality assurance are real. The reviewer-side, author-side, and area-chair-side applications in §\ref{sec:applications} are diagnostic decision-support layers; they do not write text on behalf of users. The released corpus is not intended for reviewer profiling, for automated decision-making in peer-review venues, or for training general-purpose language models.

\section*{Acknowledgments}
We thank the four additional annotators for their voluntary contributions to the development-set validation.

\paragraph{AI assistance.}
The authors used ChatGPT (OpenAI) to polish and rephrase their notes and drafts; some passages were drafted from those notes. AI assistants also supported literature discovery and source checks. Building on the authors' analysis code, Codex (OpenAI) assisted with analysis and reproduction code and numerical checks. Claude (Anthropic) provided separate checks by rerunning reproduction code and comparing reported results, tables, and cross-references against the artifacts. The authors developed the research questions, framework, and annotation guideline, made the final methodological and interpretive decisions, and reviewed and take responsibility for all content.

\bibliography{anthology,custom}


\clearpage

\appendix

\section{Annotation Guideline for ReviewMap}
\label{app:guideline}

This appendix reproduces the annotation guideline used in the experiments (§\ref{sec:method:pipeline}, §\ref{sec:results}). It accompanies the system and output-format instructions in Appendix~\ref{app:actual-prompt}. The guideline refers generally to reading the abstract, introduction, and conclusion; the experimental prompts supply the title, abstract, keywords, and review data, with claims extracted from the abstract. The term \emph{evaluative function} below denotes the \emph{evaluative acts} discussed in the main text.

\begin{tcolorbox}[breakable, enhanced,
    colback=gray!4, colframe=gray!50!black,
    boxrule=0.5pt, arc=3pt,
    left=8pt, right=8pt, top=6pt, bottom=6pt,
    fonttitle=\small\sffamily\bfseries,
    title={Annotation Guideline (released artifact)},
    coltitle=white, colbacktitle=gray!50!black]

\subsection{Purpose}

This guideline provides instructions for classifying peer-review comments from ML/CS venues by the \emph{evaluative function} each comment performs: \textbf{Assessment (ASM)} or \textbf{Calibration (CAL)}.

\begin{itemize}
    \item \textbf{ASM} evaluates \emph{what the paper does or how it does it} --- whether claims are supported and standards are met.
    \item \textbf{CAL} evaluates \emph{why the work matters} --- the contribution's novelty, significance, importance, or venue fit.
\end{itemize}

\subsection{Output format}

Each comment receives the following fields:

\begin{center}\scriptsize
\setlength{\tabcolsep}{3pt}\begin{tabular}{p{0.34\linewidth} p{0.16\linewidth} p{0.36\linewidth}}
\toprule
Field & Values & When \\
\midrule
\texttt{code} & ASM / CAL & Always \\
\texttt{claim\_relevant} & yes / no & ASM only \\
\texttt{standard\_relevant} & yes / no & ASM only \\
\texttt{grounded} & yes / no & CAL only \\
\texttt{note} & free text & When deliberating between codes, or when recording grounding evidence \\
\bottomrule
\end{tabular}
\end{center}

\paragraph{Coding unit.} Each individual comment as delimited by the reviewer under Strengths (S1, S2, \dots), Weaknesses (W1, W2, \dots), or Questions (Q1, Q2, \dots). Code only official reviews.

\subsection{Identifying claims (required before coding)}

Before coding any comments, read the paper's \textbf{abstract}, \textbf{introduction}, and \textbf{conclusion} and list the main claims. A claim is any assertion the paper explicitly makes --- empirical results, scope, design properties, or performance characteristics.

Examples of claims:
\begin{itemize}
    \item ``achieves state-of-the-art performance on X''
    \item ``our method is model-agnostic and can be applied to any architecture''
\end{itemize}

This claim list is your reference for determining \texttt{claim\_relevant} below.

\subsection{Category definitions}

\subsubsection{Assessment (ASM)}

A comment is ASM when it evaluates \emph{what the paper does or how it does it} --- whether claims are supported, methods are sound, standards are met, or presentation is adequate. ASM examines the paper's content against its own assertions or against accepted scholarly practices.

To recognize ASM, ask two questions:
\begin{itemize}
    \item Does this comment evaluate whether a \emph{specific claim} is supported by evidence? (paper vs.\ its own assertions)
    \item Does this comment evaluate whether the paper meets a \emph{field-wide standard}? (paper vs.\ accepted practices)
\end{itemize}
If either answer is yes, the comment is ASM.

\emph{ASM encompasses both positive and negative assessments:}
\begin{itemize}
    \item ``The analysis clearly demonstrates the contribution of each proposed component.'' --- evaluates what the paper did (claim support).
    \item ``The paper claims real-time performance but the reported latency excludes preprocessing time.'' --- evaluates what the paper claimed vs.\ showed.
    \item ``The paper is well-written and easy to follow.'' --- evaluates how the paper presents its work.
    \item ``The experimental setup is not described in sufficient detail for reproduction.'' --- evaluates what is missing (standard).
\end{itemize}

\subsubsection{Calibration (CAL)}

A comment is CAL when it evaluates \emph{why the work matters} --- positioning the contribution in the field by judging its novelty, significance, importance, or venue fit.

\emph{CAL encompasses both positive and negative calibrations:}
\begin{itemize}
    \item ``This is the first computational work to link X and Y.'' --- positions the work as novel; the point is why it matters, not what was done.
    \item ``The novelty is limited --- similar approaches have been widely explored.'' --- judges why the work does not matter (novelty).
    \item ``This will change how the field thinks about X.'' --- judges why the work matters (significance).
    \item ``The contribution is incremental.'' --- judges why the work does not matter (significance).
\end{itemize}

\subsection{Distinguishing ASM vs.\ CAL}

When a comment touches both assessment and calibration, code the \emph{dominant evaluative function}. To determine dominance, ask: is the comment ultimately about \emph{what the paper does}, or about \emph{why the work matters}?

Use the \texttt{note} field to record any secondary dimension you considered. Interrogative form does not determine the label --- questions should be coded by the evaluative function they perform.

\begin{center}\footnotesize
\setlength{\tabcolsep}{3pt}\begin{tabular}{p{0.38\linewidth} c p{0.38\linewidth}}
\toprule
Comment & Code & Reasoning \\
\midrule
``This is an invaluable contribution with careful detail and rigor'' & CAL & Mixed: ultimately about why the work matters (invaluable), not about the detail itself \\
``The results are impressive but the comparison with prior work is unfair'' & ASM & Mixed: ultimately about what the paper does (unfair comparison), not about impressiveness \\
\bottomrule
\end{tabular}
\end{center}

\subsection{ASM attributes}

For each comment coded ASM, record two attributes:

\paragraph{\texttt{claim\_relevant}.}
Does the comment evaluate the support for a \emph{specific claim} listed above? (yes / no). This checks the paper against \emph{its own assertions} --- did the paper do what it said?

\paragraph{\texttt{standard\_relevant}.}
Does the comment evaluate whether the paper meets a \emph{field-wide standard} that applies regardless of what the paper claims? (yes / no). This checks the paper against \emph{accepted scholarly practices} --- did the paper do what is expected?

Both can be yes simultaneously. At least one must be yes; if neither applies, reconsider whether the comment is CAL.

\subsection{CAL attribute}

For each comment coded CAL, record:

\paragraph{\texttt{grounded}.}
Does the review text contain \emph{traceable evidence} supporting the value judgment? (yes / no). If yes, briefly note the evidence in the \texttt{note} field.

Traceable evidence includes the reviewer's own assessment of the paper, a specific reference to the paper's content, or a comparison with other work. Identifying the feature being evaluated does not by itself count as evidence.

\begin{itemize}
    \item Grounded: ``The improvement is marginal --- only X\% over the compared method.'' (evidence: inline, X\% figure)
    \item Ungrounded: ``This is a strong contribution to the field.''
\end{itemize}

\subsection{Uncertain cases (FLAG)}

If you cannot determine whether a comment is ASM or CAL, output FLAG as the code.

\end{tcolorbox}

\raggedbottom 
\section{Annotation Prompt and Worked Examples}
\label{app:actual-prompt}

The template below specifies the system instructions, input fields, claim-extraction task, and output format. The guideline placeholder is filled with the text in Appendix~\ref{app:guideline}. Each prompt groups a paper's review comments under reviewer score and confidence headers. Paper content is limited to the title, abstract, and keywords, and claims are extracted from the abstract. For grounding, the wrapper refers to a specific assessment; the accompanying guideline gives the broader traceable-evidence examples. The template preserves both formulations used in the experiment. Appendix~\ref{app:guideline}'s act table lists ASM/CAL; its fallback rule and the wrapper additionally permit FLAG when the act cannot be determined (Figure~\ref{fig:prompt-template}).

The ASM example in §\ref{sec:framework} is paper \texttt{6wA4qpyyU9}, reviewer \texttt{CW52}, comment S2; the CAL example is paper \texttt{ZfCCwJ4Wcs}, reviewer \texttt{jvoP}, comment S2. Quotations omit list markers and terminal punctuation. Their final labels are ASM/P1=yes/P2=no and CAL/grounded=no, respectively, with three-model agreement on each applicable field. The output format assigns one row to each expected comment; aggregated records also include vote counts and unresolved states.

\paragraph{Human reading materials.}
The archived human sheets contain the title, abstract, sectioned review comments, and reviewer scores and confidence, without full-paper attachments. The four additional annotators' cover note specifies Level~1 judgments from the abstract and review context. Their sheets omit the keywords, abstract claim-extraction task, and Level~2 fields present for the author anchor and model prompts. All use the shared guideline. Although it refers to the introduction and conclusion, these were not supplied; external consultation was not recorded.

\paragraph{Populated input/output example.}
The following excerpt pairs the CAL example above with its actual Qwen run~1 output. We omit the system message, paper context, other comments, generated preamble and claims, and other output rows. The note is from this individual response, not a consensus explanation. Full populated requests and parsed run outputs accompany the artifact.

\noindent\begin{minipage}{\columnwidth}
\textbf{Input excerpt}
\begin{Verbatim}[fontsize=\footnotesize,breaklines=true,breakanywhere=false,breaksymbolleft={},breaksymbolright={},frame=single,framerule=0.3pt,rulecolor=\color{ReviewMapFrameGray},framesep=5pt]
### Reviewer_jvoP (score: 8, confidence: 4)

| comment_id | section | text |
|------------|---------|------|
| jvoP_S2 | Strength | "- The idea is very simple but impactful." |
\end{Verbatim}
\noindent\textbf{Generated output excerpt}
\begin{Verbatim}[fontsize=\footnotesize,breaklines=true,breakanywhere=false,breakafter={|},breakaftersymbolpre={},breakaftersymbolpost={},breaksymbolleft={},breaksymbolright={},frame=single,framerule=0.3pt,rulecolor=\color{ReviewMapFrameGray},framesep=5pt]
| comment_id | code | claim_relevant | standard_relevant | grounded | note |
|------------|------|----------------|-------------------|----------|------|
| jvoP_S2 | CAL | NA | NA | no | Judgment of impact without evidence |
\end{Verbatim}
The final consensus agrees on CAL/grounded=no; raw \texttt{NA} becomes \texttt{not\_applicable} in the aggregate schema.
\end{minipage}

\begin{figure*}[p]
\subsection*{System message}
\begin{Verbatim}[fontsize=\footnotesize,breaklines=true,breakanywhere=false,breaksymbolleft={},breaksymbolright={},frame=single,framerule=0.3pt,rulecolor=\color{ReviewMapFrameGray},framesep=7pt]
You are an annotation assistant for a peer review analysis study. Your task is to classify each review comment according to the coding scheme described in the Annotation Guideline below.

The framework distinguishes between:
- ASM (Assessment): comments that evaluate whether something in the paper is true, adequate, or met
- CAL (Calibration): comments that calibrate the contribution's novelty, significance, or venue fit

ASM and CAL are the two coding categories. Use FLAG only if genuinely uncertain.

IMPORTANT: Output ONLY the requested output. Do not include any thinking, reasoning, or explanation.

<annotation_guideline>
{FULL ANNOTATION GUIDELINE}
</annotation_guideline>
\end{Verbatim}

\subsection*{User message}
\begin{Verbatim}[fontsize=\footnotesize,breaklines=true,breakanywhere=false,breaksymbolleft={},breaksymbolright={},frame=single,framerule=0.3pt,rulecolor=\color{ReviewMapFrameGray},framesep=7pt]
## Paper Information

**Title:** {TITLE}
**Abstract:** {ABSTRACT}
**Keywords:** {KEYWORDS}

## Step 1: Identify the paper's main claims

Read the abstract above and list the paper's main claims (2-4 claims). Extract each claim as a direct quote from the abstract — do not paraphrase or rewrite. A "claim" is any property the paper explicitly asserts — including descriptive assertions, not only experimental results.

## Step 2: Code each comment

The comments below have already been segmented. Do NOT re-segment them. Code each comment exactly as given.

### Reviewer_R1 (score: {SCORE}, confidence: {CONFIDENCE})

| comment_id | section | text |
|------------|---------|------|
| R1_{COMMENT_ID} | {SECTION} | "{COMMENT_TEXT}" |

## Output Format

### Claims
1. [claim 1]
2. [claim 2]
...

### Coding Results

| comment_id | code | claim_relevant | standard_relevant | grounded | note |
|------------|------|----------------|-------------------|----------|------|
Rules for the table:
- comment_id: use exactly the IDs from the input table above
- code: ASM, CAL, or FLAG (if genuinely uncertain)
- claim_relevant: yes or no if code is ASM (does it evaluate a specific claim?), otherwise NA
- standard_relevant: yes or no if code is ASM (does it evaluate a field-wide standard?), otherwise NA
- grounded: yes or no if code is CAL (does the calibration reference a specific assessment?), otherwise NA
- note: brief explanation only if you deliberated between codes, otherwise leave blank

[Repeat reviewer headers and comment rows as in the source input.]
\end{Verbatim}
\captionsetup{skip=5pt}
\caption{Annotation prompt template. Placeholders are filled from the released inputs and annotation guideline.}
\label{fig:prompt-template}
\end{figure*}


\section{Reproducibility}
\label{app:reproducibility}

\paragraph{Implementation.}
The three open-weight models were served with vLLM (BF16), with temperature $0.6$. The provenance record lists output limits of Qwen $16$K, Nemotron $20$K, and Gemma $4$K; these are not a complete history of the requests that produced the main annotations. Recovery records document $14$K retries for $57$ Qwen, $43$ Nemotron, and $4$ Gemma paper--runs, followed by $20$K retries for $10$ of those Nemotron paper--runs. Complete initial per-request logs are unavailable. Available server logs record top-$p=0.95$ for all three, top-$k=20$ for Qwen, and $64$ for Gemma; other sampling settings followed the served model configurations where documented. No hyperparameter search was performed. Archived prompts, parsed outputs, and analysis code support deterministic reproduction of aggregation, not identical fresh inference.

\paragraph{Hardware and compute budget.}
Annotation used two hardware configurations: a workstation with dual NVIDIA RTX Pro 6000 Blackwell GPUs ($96$~GB each), with Qwen and Gemma on one GPU each, and a server with dual NVIDIA B200 GPUs ($192$~GB each), with Nemotron at tensor-parallel TP=2. Table~\ref{tab:gpu-hours} reports GPU-hours, including both GPUs for Nemotron. The 3-model $\times$ 3-run $\times$ 500-paper annotation required approximately $352$ GPU-hours, or $42.4$ GPU-minutes per paper across nine model passes. Qwen timings were measured from per-run logs. Gemma and Nemotron timings were estimated from truncation-retry samples at the production output limits because complete timing logs were unavailable.

\begin{table*}[t]
\centering\footnotesize
\setlength{\tabcolsep}{6pt}
\begin{tabular}{l l c r r r}
\toprule
Model (BF16) & Hardware & GPUs & \makecell{Per single pass\\(GPU-min)} & \makecell{Per paper (3 runs)\\(GPU-min)} & \makecell{Full pipeline\\(GPU-h)} \\
\midrule
Qwen 3.5--27B & RTX Pro 6000 (96~GB) & 1 & 8.5 & 25.5 & \textbf{212.3} \\
Gemma 4--31B & RTX Pro 6000 (96~GB) & 1 & 0.7 & 2.0 & \textbf{16.3} \\
Nemotron 3-Super 120B & NVIDIA B200 (192~GB) & 2 & 5.0 & 14.9 & \textbf{124.0} \\
\midrule
\textbf{Total} (all 3 models) & two hardware configurations & --- & \textbf{14.2} & \textbf{42.4} & \textbf{$\approx 352$} \\
\bottomrule
\end{tabular}
\caption{Compute budget for the main-set annotation. \emph{Per single pass} = one paper through one model in one SC run; \emph{per paper (3 runs)} = one paper through all three SC runs of one model; \emph{full pipeline} = the entire 3-model $\times$ 3-run $\times$ 500-paper annotation that produced the released artifact. Nemotron's TP=2 contributes two GPU-hours per elapsed hour.}
\label{tab:gpu-hours}
\end{table*}

\paragraph{Sampling and reproduction.}
Corpus sampling (§\ref{sec:method:corpus}) uses \texttt{SEED=42}. Reproduction starts from the released, fixed segmented comments. Aggregation deterministically maps the archived stochastic outputs to the reported labels, using fixed voter orders for act ties and explicit attribute abstention. The artifact includes both the run outputs and aggregated labels; fresh inference may produce different annotations.

\paragraph{Data and analysis reproduction.}
\texttt{Data/main/aggregation/} contains final labels and summary statistics; \texttt{Data/main/runs/} contains the three parsed runs per model. The corpus and all 512 main/dev prompts are accompanied by an input specification and checksums. \texttt{Data/dev/validation/} contains anonymized H1--H5 act labels, dev model outputs, and the annotation protocol; H1 denotes the author anchor. Original final act labels are retained in \texttt{Data/reference/} to check preservation under reaggregation. Per-model aggregate files are regenerated from the archived runs.

The artifact README provides one command for deterministic aggregation, development-set validation, and the corpus and sensitivity analyses. These calculations run without a GPU or network using Python 3.10 or newer and its standard library. The software also includes prompt reconstruction and optional plot rendering with matplotlib. A checksum manifest covers the released files.

\input{appendices/evidence_tables.tex}
\clearpage 
\twocolumn 
\flushbottom


\section{ReviewMap as Reviewer Guideline}
\label{app:reviewer-guide}

This appendix adapts the released \emph{Practical Reviewer Guideline} (§\ref{sec:apps:reviewer}), which originated in peer-review teaching and informed the annotation scheme. It organizes human reviewing around assessment and calibration: both involve judgment, directed at different evaluative targets. Appendix~\ref{app:guideline} presents the annotation instrument used in the experiments.

\begin{tcolorbox}[breakable, enhanced,
    colback=gray!4, colframe=gray!50!black,
    boxrule=0.5pt, arc=3pt,
    left=8pt, right=8pt, top=6pt, bottom=6pt,
    fonttitle=\small\sffamily\bfseries,
    title={Practical Reviewer Guideline},
    coltitle=white, colbacktitle=gray!50!black]

\subsection{Before you write: reading the paper}

Anchor the review in the paper's stated aims, claims, and evidence. Before writing, identify the following:
\begin{itemize}
    \item \emph{What does the paper claim?} Read abstract, introduction, and conclusion carefully. List the specific claims the authors make.
    \item \emph{What evidence does the paper provide?} For each claim, identify the corresponding evidence: experiments, proofs, ablations, visualizations, or theoretical arguments.
    \item \emph{What is the closest prior work?} Identify the most similar existing work. Understanding how the paper positions itself relative to this work is often the key to evaluating its contribution.
\end{itemize}
Use notes on these three points as the basis for the review.

\subsection{The review process: a step-by-step workflow}

The workflow below recommends independent reading and judgment. Use AI assistance only where permitted by the venue's current policy, including its confidentiality requirements.

\paragraph{Phase 1 --- Preparation (AI assistance where permitted).}
\begin{itemize}
    \item[\textbf{1.}] Read the abstract and decide whether to accept the review. Decline if outside your area of competence.
    \item[\textbf{2.}] Build your reference landscape. List cited related works; where venue policy permits, use AI to search for additional recent works.
    \item[\textbf{3.}] Familiarize yourself with key references, optionally using AI assistance where permitted.
    \item[\textbf{4.}] Prepare comparison points: what each related work does differently, where claims overlap, how the proposed method should be positioned.
\end{itemize}

\paragraph{Phase 2 --- Reading the paper independently.}
\begin{itemize}
    \item[\textbf{5.}] Read the paper yourself, with your prepared comparison points. Independent reading helps reviewers assess support for the paper's claims and scholarly standards, then judge the contribution's significance.
\end{itemize}

\paragraph{Phase 3 --- Post-reading verification.}
\begin{itemize}
    \item[\textbf{6.}] Check for additional missing references. Your understanding after reading the paper allows precise queries.
\end{itemize}

\paragraph{Phase 4 --- Writing and finalizing.}
\begin{itemize}
    \item[\textbf{7.}] Write the review in your own words.
    \item[\textbf{8.}] Where the venue permits it, optionally use AI to check grammar, clarity, and tone. Retain responsibility for the review's judgments and recommendation.
    \item[\textbf{9.}] Submit.
\end{itemize}

\subsection{Structure of a review}

A well-structured review serves authors, the area chair, and other reviewers:
\begin{itemize}
    \item \textbf{Summary (1 paragraph).} What the paper does in your own words: problem, approach, key results.
    \item \textbf{Strengths (2--5 points).} Specific strengths tied to concrete aspects. \emph{General:} ``The experiments are thorough.'' \emph{Specific:} ``The ablation in Table~3 isolates the contribution of each component across three datasets.''
    \item \textbf{Weaknesses, separated as major (affecting validity of core claims) vs.\ minor (presentation, secondary).} Be specific about why each issue matters.
    \item \textbf{Questions to authors.} Ask specific questions that the authors can address.
    \item \textbf{Suggestions for improvement} (optional but recommended).
    \item \textbf{Minor issues} (typos, formatting; keep brief).
\end{itemize}

\subsection{Two core principles for evaluating a paper}

The framework organizes evaluation into two stages:
\begin{enumerate}
    \item \textbf{Assessment} (this section): assess what the paper did and how it did it, using Principles~1 and 2.
    \item \textbf{Calibration} (§\ref{app:reviewer-guide:calibration}): build on these assessments to evaluate novelty and significance and formulate an overall recommendation.
\end{enumerate}
Identifying the claims and assessing their support provides a basis for evaluating novelty and significance. P1/P2 assessments address a paper's claims and shared scholarly standards; calibration addresses the contribution's value and venue fit. Reviewers can examine the evidence and criteria behind either judgment while disagreeing about their adequacy or significance.

\subsubsection{Principle 1: Claim--Evidence Alignment}

\emph{``Does this paper sufficiently support what it claims?''}

Evaluate the evidence in relation to the authors' specific claims. Steps: (1)~identify each claim (empirical, scope, design, or compliance assertions); (2)~find the corresponding evidence; (3)~assess the gap. Common patterns range from well-supported (acknowledge as strength), through overclaimed (point out the specific mismatch) and incomplete evidence (request the specific missing experiment), to no evidence (flag as major concern) and evidence contradicting the claim (point out precisely with references). \emph{Match the severity of the concern to the evidential gap.}

\subsubsection{Principle 2: Scholarly Standard Assessment}

\emph{``Does this paper meet the standards that apply regardless of what it claims?''}

Scholarly standards apply beyond a paper's explicit claims. Key areas: reproducibility; ethical considerations; fair treatment of prior work (citation accuracy, comparison fairness); accurate acknowledgment of limitations; writing quality. A paper may support its claims while falling short of these standards.

\subsubsection{How the two principles work together}

In practice, most comments fall under one principle or the other; some fall under both. \emph{Examples:} ``Claims state-of-the-art performance but does not compare with Method~X'' $\to$ Principle~1 (claim without evidence) and Principle~2 (unfair treatment of prior work); ``No code or hyperparameters provided'' $\to$ Principle~2 (scholarly standard); ``Claims robustness across diverse domains but evaluates only on one narrow dataset'' $\to$ Principle~1. The principles guide evaluation; explicit labels for individual comments are optional.

\subsection{Calibrating your judgment}
\label{app:reviewer-guide:calibration}

After completing P1/P2 assessment, evaluate the contribution's novelty and significance. \emph{Ground each judgment} by citing the P1 or P2 assessment that supports it.

\paragraph{Novelty.}
\emph{``Is this contribution new relative to existing knowledge?''} \emph{Specific:} ``This is, to my knowledge, the first work to combine X with Y.'' \emph{Supported by a citation:} ``Method~Z [Ref 12] already addresses the same problem with a similar approach, which limits the novelty of this submission.'' \emph{Unsupported:} ``The novelty is limited.''

\paragraph{Significance.}
\emph{``Does this contribution meaningfully advance the field or community?''} Novelty and significance can differ: a first solution to a problem may have limited significance, whereas a large-scale validation of a known technique may be significant. Significance includes venue-relative impact.

\paragraph{Confidence.}
State a confidence level that reflects your expertise and the evidential basis of the assessment.

\paragraph{Severity of language.}
Match the strength of your language to the strength of your evidence. A mathematical proof of an error supports direct statements; a suspicion supports a question or request for clarification; a personal preference should be identified as such.

\paragraph{Overall recommendation.}
The recommendation should \emph{follow from} P1/P2 assessments and novelty/significance judgments. If it appears inconsistent with the stated assessments (no major weaknesses but a rejection recommendation; identified fatal flaws but an acceptance recommendation), explain the reasoning or revisit the assessment.

\subsection{Common pitfalls in reviewing}

\begin{itemize}
    \item \textbf{Misidentifying the contribution type.} Evaluate a task-definition paper according to its aims and claims, rather than applying the expectations for a methods paper.
    \item \textbf{Equating simplicity with limited novelty.} Consider the potential long-term contribution of novel or interdisciplinary work. Before describing an approach as ``too simple,'' check whether the specific combination has been tried before.
    \item \textbf{Making general criticisms without supporting evidence.} ``This work is not new'' requires a reference to prior work. ``This work is not enough'' requires an explanation of the additional contribution needed and why. ``There is no [X] experiment'' requires a link to a specific claim or a relevant scholarly standard.
    \item \textbf{Omitting strengths from a critical review.} Identify the paper's strengths alongside its weaknesses, regardless of the overall recommendation.
    \item \textbf{Treating unfamiliarity as evidence of error.} Consult relevant sources or state that the aspect is outside your expertise.
    \item \textbf{Requesting experiments without a clear rationale.} Replace a general request such as ``more experiments'' with a rationale: ``an ablation comparing X vs.\ Y would clarify whether the improvement comes from component~A or B.''
    \item \textbf{Using dismissive language.} State the substantive concern directly and professionally: ``This is trivial'' $\to$ ``The technical contribution is primarily in system design''; ``The authors failed to'' $\to$ ``The paper does not currently include''.
\end{itemize}

\subsection{Pre-submission checklist}

\begin{itemize}
    \item Comprehension: does my summary accurately reflect what the paper does?
    \item Balance: have I identified both strengths and weaknesses?
    \item Specificity: is every weakness tied to a specific section, table, equation, or claim?
    \item Proportionality: does the stated severity of each concern reflect its implications for the paper?
    \item Actionability: would the authors know how to improve their paper from my review?
    \item Grounded value judgments: when I state a novelty or significance judgment, have I cited the P1 or P2 assessment that supports it?
    \item Assessment before recommendation: did I complete P1/P2 before forming my overall recommendation?
    \item Tone: would I be comfortable receiving this review for my own paper?
    \item Consistency: does my rating match the severity of the issues I raised?
    \item Confidence: does my confidence score accurately reflect my expertise on this topic?
\end{itemize}

\end{tcolorbox}

\section{Author-Side Strategy Scaffold}
\label{app:author-scaffold}

This appendix expands §\ref{sec:apps:author} with strategy definitions, the coding-to-menu mapping, venue-template adaptation, and a comparison with RbtAct's independently derived rebuttal-action taxonomy. The proposed scaffold supports authors' response planning; large-scale validation with author-side rebuttal corpora remains future work.

\subsection{Five response strategies}

We define five candidate strategies for authors planning a response and compare their scope with RbtAct's rebuttal-action taxonomy below:

\begin{itemize}
    \item \textsc{Leverage} --- cites the comment as third-party support in the rebuttal opening.
    \item \textsc{Defend} --- contests a critique using citations, data, or precedent.
    \item \textsc{Concede} --- accepts a substantive critique and offers a concrete remedy.
    \item \textsc{Accept} --- commits to a surface-level revision, such as correcting a typo or formatting, without an epistemic concession.
    \item \textsc{Clarify} --- answers directly with paper content or asks the reviewer for clarification.
\end{itemize}

\textsc{Accept} and \textsc{Concede} both involve agreement but differ in depth: \textsc{Accept} addresses presentation without revising the model or claims, while \textsc{Concede} entails a substantive update to the paper or its scope.

\subsection{Mapping codings to strategy menus}

The (act, attribute) coding provides up to five candidate strategies per comment (Table~\ref{tab:strategy-menu}); the author chooses the response. For ASM with both P1 and P2 marked yes, the menu is the union of the two rows. Unresolved attributes leave the corresponding menu branch inactive; ASM cases with both attributes marked no require direct reading. Attribute-derived options remain subject to human confirmation.

\begin{table}[h]
\centering\footnotesize
\begin{tabular}{p{0.32\columnwidth} p{0.55\columnwidth}}
\toprule
Coding & Strategy menu \\
\midrule
ASM, P1=yes & \textsc{Leverage} $\cdot$ \textsc{Defend} $\cdot$ \textsc{Concede} $\cdot$ \textsc{Clarify} \\
ASM, P2=yes & \textsc{Leverage} $\cdot$ \textsc{Defend} $\cdot$ \textsc{Concede} $\cdot$ \textsc{Clarify} $\cdot$ \textsc{Accept} \\
CAL & \textsc{Leverage} $\cdot$ \textsc{Defend} $\cdot$ \textsc{Concede} \\
FLAG & (no scaffold --- author reads directly) \\
\bottomrule
\end{tabular}
\caption{Proposed default mapping from codings to response strategies. The union of applicable rows yields at most five options. Unresolved attributes leave the corresponding row inactive; the author confirms attribute-dependent options and chooses the response.}
\label{tab:strategy-menu}
\end{table}

In the proposed interface, FLAG and three-way act disagreements prompt direct reading, and unresolved attributes retain their status. Act unanimity ($88.4\%$) and 2-of-3 agreement ($11.3\%$) appear as distinct vote states alongside the source comment. Authors confirm attribute-dependent options and can revise the coding or choose a response outside the menu.

\paragraph{How much of the mapping depends on attributes?}
Under the guideline's valid ASM combinations (P1 or P2=yes), four strategy memberships are invariant; only \textsc{Accept} depends on P2. CAL and FLAG each fix all five inclusion/exclusion decisions. Thus, for the observed act distribution, $(5\times17{,}654-14{,}203)/(5\times17{,}654)=74{,}067/88{,}270$ comment--strategy pairs ($83.9\%$) are structurally act-determined under that constraint. This counts inclusion/exclusion decisions for comment--strategy pairs. At the comment level, the act alone fixes the full menu for $3{,}451/17{,}654$ comments ($19.5\%$, CAL or FLAG), including the empty FLAG menu.

The aggregated attributes do not always satisfy that constraint: $327/14{,}203$ ASM comments ($2.3\%$) have neither P1 nor P2 marked yes ($287$ involve unresolved fields; $40$ have P1=P2=no). No ASM branch activates for these comments pending direct reading. Another $305$ have P1=yes and P2=unresolved: four options are available, while \textsc{Accept} remains pending. These counts describe the proposed mapping under the two-stage aggregation; attribute-derived options require human confirmation.

\paragraph{Relationship to DISAPERE rebuttal acts.}
DISAPERE's existing labels \citep{kennard-etal-2022-disapere} can describe realized responses to these choices: \emph{Concede criticism} corresponds to \textsc{Concede}, \emph{Answer} to \textsc{Clarify}, and dispute labels such as \emph{Reject criticism} or \emph{Mitigate criticism} to \textsc{Defend}. Its completed/promised-task labels may realize either \textsc{Accept} or \textsc{Concede}, depending on whether the change is surface-level or substantive; \emph{Accept praise} overlaps with, but does not entail, reusing praise as \textsc{Leverage}. The menu thus groups familiar response acts as choices available before drafting.

\subsection{Venue-template adaptation}

Venue-specific section templates can help narrow the candidate menu: ICLR's Strengths/Weaknesses/Questions template provides section cues (Strengths $\to$ \textsc{Leverage}; Weaknesses $\to$ \textsc{Defend}/\textsc{Concede}/\textsc{Accept}; Questions $\to$ \textsc{Clarify}); ARR's heterogeneous Comments/Suggestions/Typos sections require the coding for routing; CVPR-style rebuttals without sections rely entirely on the coding. The coding-to-menu mapping remains the venue-invariant core. Authors write the rebuttal, while the scaffold offers a paper-level overview of response options (e.g., ``7 \textsc{Leverage} candidates, 11 \textsc{Defend}/\textsc{Concede} pending, 18 \textsc{Clarify} pending'').

\subsection{Comparison with RbtAct's response taxonomy}

RbtAct \citep{wu2026rbtact} classifies observed rebuttal actions, while ReviewMap proposes candidate strategies before a response is written. The taxonomies partly overlap: \textsc{Defend} resembles defense without change, and revision-oriented responses can occur under \textsc{Concede} or \textsc{Accept}. RbtAct distinguishes completed revisions, specific revision plans, and vague commitments to revise; this concerns the status and specificity of a revision, rather than ReviewMap's distinction between substantive concession and presentation changes. Our menu also includes using positive comments as support (\textsc{Leverage}). These correspondences provide a conceptual comparison of the two taxonomies.

\section{Area-Chair Dashboard Example}
\label{app:dashboard}
Figure~\ref{fig:dashboard} illustrates the proposed Area-Chair Dashboard using one paper from the main corpus. The reviewer profile summarizes consensus codings, while manually annotated response links, strategies, and resolution notes illustrate how an AC could inspect the ensuing discussion.

\paragraph{From an evaluative target to a response.}
Consider LnWH/W1, ``Trivial target properties.'' The act vote is CAL~2, ASM~1. The linked response identifies the main targets as adsorption capacity and separation selectivity and points to Table~1 of the paper. We interpret this response as \textsc{Clarify}, outside the default CAL menu: it addresses the premise of the value judgment by clarifying what is evaluated. Although the CAL menu in Appendix~\ref{app:author-scaffold} lists \textsc{Leverage}/\textsc{Defend}/\textsc{Concede}, the menu remains advisory. A reviewer can inspect whether the criticism names the intended target and grounds; an author can clarify that target before defending its value; an AC can assess which part of the concern the response addresses. This retrospective, manually linked example illustrates how the proposed views organize these questions for human judgment.

\begin{figure*}[t]
\centering
\begin{tcolorbox}[colback=gray!3,colframe=gray!55!black,boxrule=0.5pt,arc=2pt,
    width=0.97\textwidth,title={ReviewMap --- discussion-phase panel view},
    fonttitle=\small\sffamily\bfseries]
\small
\textbf{Paper:} \texttt{v2oYZJ7Exo}\hfill\textbf{Recorded scores:} 2 / 4 / 6 / 8

\medskip
\textbf{Reviewer profile: counts from consensus codings}
\begin{center}
\begin{tabular}{lrrrrrrr}
\toprule
Reviewer & Score & ASM & CAL & FLAG & \multicolumn{3}{c}{CAL grounding} \\
\cmidrule(lr){6-8}
 & & & & & yes & no & unresolved \\
\midrule
G88g & 8 & 8 & 3 & 0 & 0 & 2 & 1 \\
4Vi5 & 6 & 4 & 1 & 0 & 0 & 1 & 0 \\
DEPu & 4 & 12 & 0 & 0 & 0 & 0 & 0 \\
LnWH & 2 & 8 & 4 & 0 & 1 & 2 & 1 \\
\bottomrule
\end{tabular}
\end{center}
\textbf{Inspect the unresolved cases.} LnWH/W7 and G88g/S2 each have one yes and one no grounding vote among two eligible CAL voters. Both retain an unresolved grounding label. LnWH/W1 has two yes votes among two eligible voters; its act remains contested (CAL 2, ASM 1).

\medskip
\textbf{Manually linked concern--response examples}
\begin{center}
\footnotesize
\begin{tabular}{@{}p{0.14\linewidth}p{0.24\linewidth}p{0.36\linewidth}p{0.17\linewidth}@{}}
\toprule
Comment / grounding & Review excerpt & Linked author response (paraphrase) & Strategy / resolution note \\
\midrule
LnWH/W1: yes (2/2 eligible) & ``Trivial target properties'' & R1: the main targets are adsorption capacity and separation selectivity, with non-trivial properties in Table 1. & \textsc{Clarify} (outside default menu); scope clarified \\
\addlinespace
LnWH/W7: unresolved (1 yes, 1 no) & ``Given the weak empirical gains, it's unclear whether this added complexity is warranted'' & R5: the authors report monotonic improvements, over 30\% MAE reduction, and about 30\% additional computational complexity. & \textsc{Defend}; partial \\
\bottomrule
\end{tabular}
\end{center}
\footnotesize
Resolution notes reflect our interpretation of how each response addresses the concern and do not establish reviewer acceptance. Excerpts are abridged: W7 also includes a benchmark-choice concern. Full comments and responses remain available for inspection, including concerns with unresolved grounding.

\medskip
\small
\textbf{Discussion context:} the authors publicly disavowed an unrelated third-party comment. This manually annotated discussion event appears alongside the official-review act labels.
\end{tcolorbox}
\caption{AC dashboard mockup for paper \texttt{v2oYZJ7Exo} (\S\ref{sec:apps:ac}). Scores record the post-discussion snapshot in the corpus; comment counts, act labels, grounding states, and vote counts summarize the consensus codings. Concern selection, response matching, strategy badges, resolution notes, and discussion context are manually annotated for this case study. Source evidence and uncertain labels remain visible to support the AC's judgment.}
\label{fig:dashboard}
\end{figure*}

\clearpage
\onecolumn
\twocolumn
\input{appendices/header_sensitivity}

\end{document}

%% file: authors.tex
\author{
  Kihyun Na\textsuperscript{1}\thanks{This work was done while the first author was at Handong Global University.} \and Injung Kim\textsuperscript{2}\thanks{Corresponding author: \href{mailto:ijkim@handong.edu}{ijkim@handong.edu}.} \\
  \textsuperscript{1}Department of Neurosurgery, University of Alabama at Birmingham \\
  Birmingham, AL 35294, USA \\
  \textsuperscript{2}Department of Artificial Intelligence, \\
  Computer and Electrical Engineering \\
  Handong Global University, Pohang 37554, Republic of Korea \\
  \texttt{khna@uab.edu}, \texttt{ijkim@handong.edu}
}

%% file: appendices/evidence_tables.tex
\section{Agreement and Validation Details}
\label{app:evidence}

\paragraph{Aggregation conventions.}
All main-set calculations use the $17{,}654$ segmented-comment identifiers. For duplicate identifiers, the last row in input-file order is retained; extra identifiers are excluded and missing identifiers are recorded. Stage~1 takes the mode of observed run labels. After duplicate resolution, every retained expected row has a valid act. Stage~2 takes the mode of available valid model labels. Ties follow the first available run in order 1--3 or the model order Qwen--Nemotron--Gemma. Items with no available act remain null; all items in the main set have a model and final act label. Vote counts are retained for the $49$ three-way panel ties.

For a field conditional on ASM or CAL, a voter participates only when its act matches the selected act and its field value is yes or no. At least two matching binary votes are required independently at each stage. Consequently, a Stage-1 unresolved field supplies no binary vote at Stage~2. Unresolved records either a binary tie or insufficient valid votes; not applicable denotes a field outside the selected act. Neither state is counted as no. The released records preserve the support counts and reasons.

The guideline requires at least one ASM field to be yes, but independent field voting can yield P1=P2=no. This occurs for $126$, $79$, and $1$ comments in Qwen, Gemma, and Nemotron, respectively, and $40$ in the final consensus. These labels are retained and reported as violations of the joint constraint, separately from unresolved fields.

\paragraph{Conditional agreement.}
Table~\ref{tab:full-agreement} distinguishes within-model Fleiss' $\kappa$ (three raw runs) from between-model Cohen's $\kappa$ (two emitted model labels). For acts, $N_{\mathrm{elig}}=17{,}654$ and $n$ contains only complete valid ratings. For attributes, $N_{\mathrm{elig}}$ contains items whose compared acts all match the applicable act; $n$ further requires complete binary field values. Coverage is $n/N_{\mathrm{elig}}$, and observed agreement is pairwise agreement (averaged over rater pairs for Fleiss). The artifact also reports $n/17{,}654$, undefined-coefficient reasons, and three-model Fleiss coefficients. These are measures of label consistency under the annotation protocol.

\begin{table*}[t]
\centering\footnotesize
\setlength{\tabcolsep}{6pt}
\begin{tabular}{llrrrrr}
\toprule
Model / pair & Field & $\kappa$ & $n$ & $N_{\mathrm{elig}}$ & Coverage (\%) & Agreement \\
\midrule
\multicolumn{7}{l}{\emph{Intra-model Fleiss' $\kappa$ over three raw runs}}\\
Qwen & Act & 0.842 & 17,654 & 17,654 & 100.00 & 0.950 \\
Qwen & P1 & 0.630 & 13,532 & 13,613 & 99.40 & 0.815 \\
Qwen & P2 & 0.563 & 13,350 & 13,613 & 98.07 & 0.827 \\
Qwen & Grounded & 0.546 & 2,607 & 2,609 & 99.92 & 0.905 \\
Gemma & Act & 0.959 & 17,653 & 17,654 & 99.99 & 0.987 \\
Gemma & P1 & 0.891 & 13,156 & 13,945 & 94.34 & 0.946 \\
Gemma & P2 & 0.860 & 13,652 & 13,945 & 97.90 & 0.936 \\
Gemma & Grounded & 0.864 & 2,939 & 2,946 & 99.76 & 0.934 \\
Nemotron & Act & 0.780 & 17,642 & 17,654 & 99.93 & 0.921 \\
Nemotron & P1 & 0.581 & 12,589 & 12,686 & 99.24 & 0.791 \\
Nemotron & P2 & 0.456 & 12,643 & 12,686 & 99.66 & 0.821 \\
Nemotron & Grounded & 0.587 & 2,509 & 2,509 & 100.00 & 0.794 \\
\midrule
\multicolumn{7}{l}{\emph{Inter-model Cohen's $\kappa$ over emitted model labels}}\\
Q--N & Act & 0.767 & 17,654 & 17,654 & 100.00 & 0.923 \\
Q--N & P1 & 0.588 & 13,107 & 13,435 & 97.56 & 0.794 \\
Q--N & P2 & 0.481 & 13,005 & 13,435 & 96.80 & 0.821 \\
Q--N & Grounded & 0.212 & 2,512 & 2,625 & 95.70 & 0.615 \\
Q--G & Act & 0.772 & 17,654 & 17,654 & 100.00 & 0.927 \\
Q--G & P1 & 0.583 & 13,050 & 13,577 & 96.12 & 0.791 \\
Q--G & P2 & 0.504 & 13,210 & 13,577 & 97.30 & 0.789 \\
Q--G & Grounded & 0.293 & 2,521 & 2,557 & 98.59 & 0.705 \\
N--G & Act & 0.753 & 17,654 & 17,654 & 100.00 & 0.916 \\
N--G & P1 & 0.560 & 12,600 & 13,274 & 94.92 & 0.780 \\
N--G & P2 & 0.407 & 12,888 & 13,274 & 97.09 & 0.762 \\
N--G & Grounded & 0.480 & 2,440 & 2,529 & 96.48 & 0.739 \\
\bottomrule
\end{tabular}
\caption{Main-set agreement with contributing and act-eligible denominators. Q, G, and N denote Qwen, Gemma, and Nemotron. Each attribute comparison conditions on its own applicable act and complete yes/no ratings.}
\label{tab:full-agreement}
\end{table*}

\paragraph{Matched-item comparison.}
Within- and between-model coefficients need not use identical items in Table~\ref{tab:full-agreement}. Table~\ref{tab:matched-agreement} therefore restricts each model-pair/field comparison to items with all six valid raw ratings and both valid model labels, with every act applicable for attributes. On grounding, each model's intra-$\kappa$ exceeds the pair's inter-$\kappa$ on the same items. This ordering is not universal for P1/P2 and does not identify a causal source of model disagreement.

\begin{table*}[t]
\centering\footnotesize
\begin{tabular}{llrrrr}
\toprule
Pair A--B & Field & Matched $n$ & Intra A & Intra B & Inter A--B \\
\midrule
Q--N & Act & 17,642 & 0.842 & 0.780 & 0.768 \\
Q--N & P1 & 12,038 & 0.639 & 0.578 & 0.580 \\
Q--N & P2 & 11,925 & 0.563 & 0.454 & 0.475 \\
Q--N & Grounded & 2,117 & 0.554 & 0.567 & 0.208 \\
Q--G & Act & 17,653 & 0.842 & 0.959 & 0.772 \\
Q--G & P1 & 12,266 & 0.629 & 0.890 & 0.582 \\
Q--G & P2 & 12,563 & 0.562 & 0.857 & 0.497 \\
Q--G & Grounded & 2,257 & 0.547 & 0.867 & 0.280 \\
N--G & Act & 17,641 & 0.780 & 0.959 & 0.753 \\
N--G & P1 & 11,476 & 0.579 & 0.893 & 0.556 \\
N--G & P2 & 12,016 & 0.455 & 0.860 & 0.407 \\
N--G & Grounded & 2,159 & 0.578 & 0.866 & 0.475 \\
\bottomrule
\end{tabular}
\caption{Matched-item sensitivity: intra-model Fleiss' $\kappa$ and inter-model Cohen's $\kappa$ on identical items within each row. The artifact includes observed agreement for each coefficient.}
\label{tab:matched-agreement}
\end{table*}

\begin{table*}[t]
\centering\footnotesize
\begin{tabular}{lrrrrrrrr}
\toprule
Field & Applicable & Yes & No & Unresolved & 3/3 & 2/3 & 2/2 & Coverage (\%) \\
\midrule
P1 & 14,203 & 6,451 & 7,221 & 531 & 8,445 & 3,905 & 1,322 & 96.26 \\
P2 & 14,203 & 10,223 & 3,405 & 575 & 8,646 & 3,894 & 1,088 & 95.95 \\
Grounded & 3,067 & 1,933 & 871 & 263 & 1,160 & 1,083 & 561 & 91.42 \\
\bottomrule
\end{tabular}
\caption{Final attribute counts. Coverage is the proportion of applicable fields resolved as yes or no. Unresolved P1/P2/grounding fields comprise $385/395/183$ binary ties and $146/180/80$ cases with insufficient support, respectively.}
\label{tab:full-attribute-coverage}
\end{table*}

Table~\ref{tab:full-attribute-coverage} reports final field counts, support tiers, and resolved coverage.

\paragraph{Output quality.}
Table~\ref{tab:raw-audit} breaks down Table~\ref{tab:failure-modes} by run, each with $17{,}654$ expected comments. Extra identifiers are counted once per run; duplicates count additional rows. Applicable-field NAs are counted after retaining the last duplicate row. All retained expected rows have valid acts. P1=P2=no outputs are reported separately as violations of the ASM constraint. Unresolved aggregation results are recorded separately from generation errors.

For example, Qwen run~3 adds Q2 for paper \texttt{32QQlzm9ft}, reviewer \texttt{cpQL}, whose input IDs are S1, W1--W3, and Q1. The extra row (ASM; P1=yes) is excluded because its identifier is outside the input set. Input Q1 already contains two questions; this illustrates an extra identifier, not evidence of invented review content.

\begin{table*}[t]
\centering\footnotesize
\setlength{\tabcolsep}{4pt}
\begin{tabular}{lrrrrrrr}
\toprule
Model / run & Extra & Missing & Dup. & \makecell{P1\\NA / applicable} & \makecell{P2\\NA / applicable} & \makecell{Grounded\\NA / applicable} & Both-no \\
\midrule
Qwen / 1 & 0 & 0 & 60 & 34 / 14,289 & 182 / 14,289 & 0 / 3,113 & 175 \\
Qwen / 2 & 0 & 0 & 50 & 48 / 14,292 & 94 / 14,292 & 0 / 3,147 & 166 \\
Qwen / 3 & 1 & 0 & 4 & 2 / 14,248 & 37 / 14,248 & 2 / 3,185 & 134 \\
Gemma / 1 & 44 & 1 & 0 & 418 / 14,119 & 166 / 14,119 & 4 / 3,110 & 60 \\
Gemma / 2 & 52 & 0 & 6 & 628 / 14,082 & 188 / 14,082 & 6 / 3,107 & 96 \\
Gemma / 3 & 49 & 0 & 8 & 408 / 14,109 & 166 / 14,109 & 1 / 3,079 & 103 \\
Nemotron / 1 & 9 & 3 & 0 & 37 / 13,773 & 3 / 13,773 & 1 / 3,243 & 0 \\
Nemotron / 2 & 8 & 0 & 18 & 70 / 13,723 & 39 / 13,723 & 0 / 3,288 & 0 \\
Nemotron / 3 & 14 & 10 & 0 & 15 / 13,749 & 4 / 13,749 & 0 / 3,309 & 2 \\
\bottomrule
\end{tabular}
\caption{Per-run main-set output audit. Applicable denominators use each retained raw row's act, so they differ from final-consensus denominators.}
\label{tab:raw-audit}
\end{table*}

\paragraph{Five-human and model validation.}
Each of the five annotators labels all 470 dev comments ($2{,}350$ labels). The human-reference distribution is ASM $406$, CAL $51$, FLAG $13$. All five agree on $352$ items ($74.9\%$), and at least four agree on $428$ ($91.1\%$). Of the $118$ contested items, $102$ involve only ASM/CAL disagreement. The human tie rule resolves one item; the individual annotations are retained without adjudication.

Table~\ref{tab:dev-model-details} compares each model's run-aggregated labels and the two panels with the five-human reference. Table~\ref{tab:dev-panel-confusion} reports the corresponding confusion matrices. Using the author anchor alone as reference yields Cohen's $\kappa=.620$ with the open-weight panel and $.653$ with the frontier panel. Across individual humans, the open-weight comparisons range from $.505$ to $.620$ (mean $.569$), and frontier comparisons from $.518$ to $.658$ (mean $.593$). Pooling these ten comparisons gives $.505$--$.658$ (mean $.581$), overlapping the ten human--human comparisons ($.468$--$.649$, mean $.555$), without establishing statistical equivalence.

For context, DISAPERE reports Cohen's $\kappa=.447$--$.605$ across six sentence-annotation tasks \citep{kennard-etal-2022-disapere}. Its labels and units differ from ours, so these values do not define a performance ceiling. Inconsistency in review scores \citep{cortes2021inconsistency} concerns judgments of paper merit, distinct from agreement on the act expressed in a comment.

\begin{table*}[t]
\centering\footnotesize
\begin{tabular}{lrrrrrrr}
\toprule
Model / panel & $\kappa$ & Agreement (\%) & Macro F1 & CAL count & CAL precision & CAL recall & CAL F1 \\
\midrule
Qwen & 0.616 & 89.57 & 0.665 & 80 & 0.575 & 0.902 & 0.702 \\
Gemma & 0.630 & 89.57 & 0.830 & 76 & 0.526 & 0.784 & 0.630 \\
Nemotron & 0.602 & 88.94 & 0.751 & 74 & 0.527 & 0.765 & 0.624 \\
GPT-5.5 & 0.660 & 90.64 & 0.752 & 79 & 0.582 & 0.902 & 0.708 \\
Gemini 3.1 Pro & 0.690 & 91.70 & 0.787 & 73 & 0.630 & 0.902 & 0.742 \\
Opus 4.6 & 0.645 & 90.85 & 0.751 & 68 & 0.618 & 0.824 & 0.706 \\
Open-weight panel & 0.633 & 90.00 & 0.779 & 75 & 0.560 & 0.824 & 0.667 \\
Frontier panel & 0.686 & 91.49 & 0.783 & 75 & 0.613 & 0.902 & 0.730 \\
\bottomrule
\end{tabular}
\caption{Dev comparisons against the five-human mode ($n=470$ throughout; human-reference CAL count $=51$). A panel does not necessarily outperform every constituent model.}
\label{tab:dev-model-details}
\end{table*}

\input{appendices/dev_panel_tables.tex}

\input{appendices/dev_section_diagnostics.tex}

\paragraph{Model disagreement as an inspection cue.}
Table~\ref{tab:dev-vote-inspection} groups the same 470 comments by agreement among each panel's three run-voted model labels. The non-unanimous subsets include two open-weight and one frontier three-way ties, retained under the documented panel tie-break. This is a post-hoc descriptive analysis of 12 development papers, with different class compositions across subsets and an imperfect human reference. For the open-weight panel, the 69 non-unanimous items contain 30 of 47 human-reference mismatches; 17 remain among unanimous items. The association motivates displaying dissent for closer reading, without treating unanimity as a reason to bypass inspection. It does not measure effects on human decisions or review time.

\input{appendices/area_calibration.tex}

\paragraph{Dev aggregation.}
All six models supply three act votes per comment except Gemma, whose runs~1 and~3 each omit two comments; these use the available run~2 label. Two Nemotron run ties follow the first-available-run rule. The open-weight and frontier panels have two and one three-way ties, respectively, resolved by their specified model orders. Reaggregation from archived runs reproduces all model and panel acts. Human validation covers Level~1 only.

\begin{table*}[t]
\centering\footnotesize
\begin{tabular}{lrrr}
\toprule
Audit count & Qwen & Gemma & Nemotron \\
\midrule
\multicolumn{4}{l}{\emph{Raw outputs: counts summed over three runs}}\\
Extra unique IDs (outside expected set) & 1 & 145 & 31 \\
Missing expected IDs & 0 & 1 & 13 \\
Duplicate rows & 114 & 14 & 18 \\
NA in applicable P1 / P2 / grounded fields & 84 / 313 / 2 & 1{,}454 / 520 / 11 & 122 / 46 / 1 \\
\midrule
\multicolumn{4}{l}{\emph{Stage-1 aggregation: unresolved attribute fields in emitted model labels}}\\
P1 / P2 / grounded & 130 / 274 / 53 & 460 / 185 / 16 & 293 / 295 / 152 \\
\bottomrule
\end{tabular}
\caption{Raw-output quality and aggregation coverage have different units. Raw counts are summed across $3\times17{,}654$ expected comment--run opportunities per model (an extra ID can recur across runs); raw attribute-field counts follow last-row duplicate resolution. Stage-1 unresolved counts refer to one emitted label per expected comment and are deliberate abstentions, not raw-generation errors. Appendix~\ref{app:evidence} gives per-run and coverage denominators.}
\label{tab:failure-modes}
\end{table*}

%% file: appendices/dev_panel_tables.tex
\begin{table*}[t]
\begin{minipage}[t]{0.48\textwidth}
\vspace{0pt}
\centering\footnotesize
\setlength{\tabcolsep}{4pt}\begin{tabular}{llrrr}
\toprule
Panel & Human reference & ASM & CAL & FLAG \\
\midrule
Open-weight & ASM & 373 & 32 & 1 \\
Open-weight & CAL & 9 & 42 & 0 \\
Open-weight & FLAG & 4 & 1 & 8 \\
Frontier & ASM & 377 & 28 & 1 \\
Frontier & CAL & 5 & 46 & 0 \\
Frontier & FLAG & 5 & 1 & 7 \\
\bottomrule
\end{tabular}
\caption{Dev panel confusion matrices. Rows are the five-human mode; columns are panel predictions.}
\label{tab:dev-panel-confusion}
\end{minipage}\hfill
\begin{minipage}[t]{0.48\textwidth}
\vspace{0pt}
\input{appendices/vote_diagnostic.tex}
\end{minipage}
\end{table*}

%% file: appendices/vote_diagnostic.tex
\centering\footnotesize
\setlength{\tabcolsep}{4pt}
\begin{tabular}{lrrr}
\toprule
Model votes & $n$ & Matches & Agreement (\%) \\
\midrule
\multicolumn{4}{l}{\emph{Open-weight panel}}\\
Unanimous & 401 & 384 & 95.8 \\
Non-unanimous & 69 & 39 & 56.5 \\
\midrule
\multicolumn{4}{l}{\emph{Frontier panel}}\\
Unanimous & 430 & 407 & 94.7 \\
Non-unanimous & 40 & 23 & 57.5 \\
\bottomrule
\end{tabular}
\caption{Dev panel agreement with the five-human mode, grouped by agreement among three per-model run-voted acts. Non-unanimous includes 2-of-3 and three-way ties; panel labels retain the specified tie-break. This groups by model votes, unlike the human-defined subsets in Table~\ref{tab:human-validation}.}
\label{tab:dev-vote-inspection}

%% file: appendices/dev_section_diagnostics.tex
\paragraph{Section-specific CAL disagreements.}
The five-human mode and open-weight panel label respectively $50/150$ and $55/150$ Strengths comments as CAL, $1/189$ and $14/189$ Weaknesses comments, and $0/131$ and $6/131$ Questions comments. Relative to the human reference, the 33 CAL false positives comprise 14 Strengths, 13 Weaknesses, and 6 Questions; all nine false negatives occur in Strengths. Fourteen of 47 reviewer--paper instances contain no human-mode CAL, versus seven under the panel, with five shared. The human reference also shows higher CAL prevalence in Strengths than Weaknesses, but only one human-mode CAL occurs in Weaknesses. This small development sample does not justify correcting the main-corpus prevalence or effect size.

%% file: appendices/area_calibration.tex
\begin{table*}[t]
\centering\footnotesize
\setlength{\tabcolsep}{3pt}
\begin{tabular}{p{0.61\textwidth}rrr}
\toprule
Primary area & Papers & CAL/all comments & CAL (\%) \\
\midrule
causal reasoning & 12 & 58/467 & 12.4 \\
learning on graphs and other geometries \& topologies & 17 & 81/630 & 12.9 \\
probabilistic methods (Bayesian methods, variational inference, sampling, UQ, etc.) & 16 & 84/547 & 15.4 \\
generative models & 37 & 203/1,256 & 16.2 \\
applications to robotics, autonomy, planning & 17 & 99/601 & 16.5 \\
transfer learning, meta learning, and lifelong learning & 18 & 104/631 & 16.5 \\
infrastructure, software libraries, hardware, systems, etc. & 11 & 53/321 & 16.5 \\
foundation or frontier models, including LLMs & 54 & 304/1,833 & 16.6 \\
applications to computer vision, audio, language, and other modalities & 48 & 258/1,546 & 16.7 \\
applications to physical sciences (physics, chemistry, biology, etc.) & 23 & 141/835 & 16.9 \\
alignment, fairness, safety, privacy, and societal considerations & 36 & 217/1,270 & 17.1 \\
other topics in machine learning (i.e., none of the above) & 21 & 134/780 & 17.2 \\
reinforcement learning & 27 & 172/987 & 17.4 \\
learning on time series and dynamical systems & 16 & 114/630 & 18.1 \\
optimization & 21 & 148/817 & 18.1 \\
applications to neuroscience \& cognitive science & 14 & 94/513 & 18.3 \\
datasets and benchmarks & 33 & 230/1,202 & 19.1 \\
neurosymbolic \& hybrid AI systems (physics-informed, logic \& formal reasoning, etc.) & 12 & 86/430 & 20.0 \\
interpretability and explainable AI & 21 & 157/783 & 20.1 \\
unsupervised, self-supervised, semi-supervised, and supervised representation learning & 29 & 223/1,082 & 20.6 \\
learning theory & 17 & 107/493 & 21.7 \\
\midrule
Total & 500 & 3,067/17,654 & 17.4 \\
\bottomrule
\end{tabular}
\caption{CAL shares across all 21 primary areas in the main corpus under the score/confidence-provided condition and two-stage aggregation. Comments are grouped by their paper's recorded primary area; each denominator includes all original comments in that area, including \textsc{Flag}. Rows are ordered by CAL share. These finite-sample descriptive rates characterize variation within ICLR 2026; they are not uncertainty bounds or evidence of cross-venue portability.}
\label{tab:area-calibration}
\end{table*}

%% file: appendices/header_sensitivity.tex
\raggedbottom
\section{Sensitivity to Score and Confidence Headers}
\label{app:header-sensitivity}

This appendix reports a supplementary input-condition comparison. The main analysis, preceding empirical tables and figures, and human and frontier-model development validation use the score/confidence-provided condition. The main compute estimate in Appendix~\ref{app:reproducibility} excludes these additional runs.

\paragraph{Comparison protocol.}
We compare the main condition with separate runs in which reviewer score and confidence headers were jointly removed. The 500 papers, 17,654 comment identifiers, guideline, and remaining prompt content are unchanged. Both conditions use three runs per model and the same two-stage aggregation rules. The removal condition uses a 16,000-token generation limit for each model; the main condition includes varying budgets and targeted retries (Appendix~\ref{app:reproducibility}). Generation settings and recovery histories were therefore not fully matched.

\paragraph{Act labels and corpus patterns.}
Final acts agree on $17{,}211/17{,}654$ comments ($97.49\%$); 443 change. Of these, 164 move from ASM to CAL and 202 from CAL to ASM; the remaining 77 involve FLAG. Overall agreement is weighted by the large ASM share. CAL remains sparse and concentrated in Strengths (Table~\ref{tab:header-sensitivity}). Both conditions have 333 zero-CAL reviewer--paper instances, but only 300 are shared: 33 leave and 33 enter this group. The zero-CAL sets therefore overlap on $300/333$ instances.

\begin{table}[!htbp]
\centering\small
\setlength{\tabcolsep}{4pt}
\begin{tabular}{@{}lrr@{}}
\toprule
Measure & Main & Removed \\
\midrule
CAL comments (\%) & 17.37 & 17.27 \\
Zero-CAL reviews (\%) & 17.03 & 17.03 \\
Strengths/Weaknesses CAL ratio & 2.819 & 2.726 \\
\midrule
Mean inter-model $\kappa$: Act & .764 & .757 \\
\quad P1 & .577 & .578 \\
\quad P2 & .464 & .450 \\
\quad Grounded & .328 & .351 \\
\bottomrule
\end{tabular}
\caption{Main condition versus score/confidence headers removed. Denominators are 17,654 comments and 1,955 review instances. Agreement values are means of three pairwise Cohen's $\kappa$ coefficients.}
\label{tab:header-sensitivity}
\end{table}

\paragraph{Agreement across annotation layers.}
The ordering of mean inter-model agreement persists. Attribute coefficients require matching applicable acts and available binary labels; samples differ by field and condition. Restricting each pair--field comparison to items eligible in both conditions also preserves this ordering. Removal-condition act coefficients range from .745 to .773.

Final attribute-state agreement is $87.38\%$ for P1 and $87.76\%$ for P2 on the 14,007 comments retaining ASM in both conditions, and $86.74\%$ for grounding on the 2,865 retaining CAL. These states include yes, no, and unresolved. Each rate uses its stated classes and conditional sample.

\paragraph{Within-condition variation.}
For a comparison at equal aggregation depth, we form three panels per condition, each using one run from each model. Mean act disagreement is $3.20\%$ across the three within-main panel pairs, $3.64\%$ across the three within-removal pairs, and $3.46\%$ across the nine between-condition pairs. These three-rating panels differ from the nine-rating, two-stage final consensus above. The between-condition differences are descriptively similar in scale to within-condition variation. The panel pairs share runs and items, so these values provide a descriptive baseline.

\paragraph{Interpretation and limits.}
The reported corpus patterns persist across the evaluated conditions, while individual labels and attribute states change. The comparison measures joint header-removal sensitivity across differing generation settings; the removal condition has no separate human validation. Review text can itself convey evaluative stance even without numeric headers. The supplementary artifact provides saved outputs, aggregation and comparison code, and a missing/duplicate-row sensitivity check; excluding all 260 comments with any missing, duplicate, or invalid act among the 18 raw ratings gives $97.52\%$ final-act agreement on 17,394 comments.